\documentclass{article}

\PassOptionsToPackage{numbers, compress}{natbib}

\usepackage[preprint]{neurips_2026}

\usepackage[utf8]{inputenc} 
\usepackage[T1]{fontenc}    
\usepackage{hyperref}       
\usepackage{url}            
\usepackage{booktabs}       
\usepackage{amsfonts}       
\usepackage{amsmath, amssymb}
\usepackage{amsthm}
\usepackage{nicefrac}       
\usepackage{microtype}      
\usepackage{xcolor}         
\usepackage{graphicx}       
\usepackage{multirow}       
\usepackage{rotating}       
\usepackage{subcaption}     

\newtheorem{theorem}{Theorem}
\newtheorem{corollary}{Corollary}
\newtheorem{lemma}{Lemma}
\newtheorem{assumption}{Assumption}

\title{\textsc{IADR}: Interface-Augmented Neural Operator for Phase-Field Mean-Curvature Flow}

\author{%
  Qinyi Zhang\thanks{Equal contribution.}\\
  School of Mathematics\\
  Sichuan University, Chengdu, China\\
  \texttt{2024322010008@stu.scu.edu.cn}\\
  \And
  Duanyu Feng\footnotemark[1]\\
  College of Computer Science\\
  Sichuan University, Chengdu, China\\
  \texttt{fengduanyuscu@stu.scu.edu.cn}\\
  \And
  Yangshuai Wang\thanks{Co-corresponding authors.}\\
  Department of Mathematics\\
  National University of Singapore\\
  \texttt{yswang@nus.edu.sg}\\
  \And
  Hao Wang\footnotemark[2]\\
  School of Mathematics\\
  Sichuan University, Chengdu, China\\
  \texttt{wangh@scu.edu.cn}\\
}

\begin{document}

\maketitle

\begin{abstract}
The Allen--Cahn equation is one of the standard phase-field models for mean-curvature flow (MCF), parameterised by the diffuse-interface width $\varepsilon$. Practical simulations span a range of $\varepsilon$, so a single model covering this range is needed. PDE-specific neural operators are accurate but must be retrained at every $\varepsilon$; general neural operators cover $\varepsilon$ in one model but lose the phase-field structure of the target. We introduce Interface-Augmented Diffusion-Reaction (\textsc{IADR}), a neural operator that addresses both shortcomings: a target network with a diffusion-and-reaction (DR) structure preserves the phase-field structure of the Allen--Cahn equation, while a hypernetwork that maps $\varepsilon$ to the weights of this target network amortises the dependence on $\varepsilon$ across the range. A single \textsc{IADR} matches the per-$\varepsilon$ specialists on in-distribution data within the prescribed range, retains their cross-shape generalisation under both $\varepsilon$- and geometry-out-of-distribution shifts, and retains their compact DR target structure with millisecond-scale inference. We further establish a conditional $K$-step rollout error bound under consistency and stability assumptions. Code is available at \url{https://github.com/ArcueidCroft/IADR}.
\end{abstract}

\section{Introduction}
\label{sec:intro}

Mean-curvature flow (MCF) is a geometric evolution law in which every point of an interface moves along its inward normal at a speed proportional to the local mean curvature, and it underlies a wide range of applications including multi-phase fluid flow and materials microstructure coarsening~\citep{bretin2022jcp, chen2002phase}. Among the numerical approaches for MCF, the phase-field approach replaces the sharp interface by a smooth transition layer of width $\varepsilon$, described by a scalar order parameter $u$. The $L^{2}$-gradient flow of the corresponding Cahn--Hilliard energy then yields the Allen--Cahn equation $\partial_t u = \Delta u - \tfrac{1}{\varepsilon^{2}} W'(u)$, the standard phase-field model of MCF and the focus of this paper. Its $\varepsilon \to 0$ limit is rigorously the sharp-interface MCF~\citep{evans1991motion, ilmanen1993convergence}, and decades of numerical analysis~\citep{chen1998applications, du2020phase} have made the Allen--Cahn equation a canonical workhorse of phase-field simulation. The interface width $\varepsilon$ is a physically meaningful tunable parameter~\citep{du2020phase}: smaller $\varepsilon$ sharpens the interface and tightens the sharp-interface limit but demands finer spatial resolution and shorter time steps, while larger $\varepsilon$ smooths the dynamics and reduces compute at the cost of geometric fidelity. Practical simulations span a range of $\varepsilon$ rather than a single fixed value, so accurate and efficient simulation across this range is a prerequisite for interface modelling.

Neural-network operators have been studied as a faster differentiable surrogate for the Allen--Cahn equation; two categories are most relevant. \emph{General operator-learning frameworks} (e.g., FNO, DeepONet) approximate the solution map of a parametric PDE family by a function-to-function network and condition on $\varepsilon$ through an input channel. \emph{PDE-specific operator learning} replaces an operator-splitting scheme such as the diffusion and reaction (DR) splitting by a trainable network with the same forward topology (a linear diffusion solve followed by a pointwise nonlinear reaction update), preserving the phase-field structure and attaining sub-grid accuracy with only a few hundred parameters per $\varepsilon$, but training a separate network for each $\varepsilon$. Both achieve useful accuracy on the Allen--Cahn equation at a fixed $\varepsilon$. General operators cover $\varepsilon$ in one model but, with no phase-field structure in the target, their behaviour on the Allen--Cahn evolution remains uncertain; PDE-specific operators retain the structure but do not share parameters across $\varepsilon$.

Inspired by DISCO's weight-injection design~\citep{morel2025disco}, we close the $\varepsilon$-generalisation gap with a single neural operator specialised to the Allen--Cahn equation, in which one part of the network encodes the phase-field structure and another part encodes the dependence on $\varepsilon$. Concretely, the \textbf{target network} retains the diffusion-and-reaction factorisation of classical operator-splitting schemes (a spatial diffusion convolution composed with a pointwise nonlinear reaction), so that the phase-field structure shared by every $\varepsilon \in E$ is hard-coded into the architecture. A separate \textbf{hypernetwork} reads the scalar $\varepsilon$ and produces the weights of this target network, so the dependence of the operator on $\varepsilon$ is concentrated in a single small MLP and shared across the range. We call the resulting operator Interface-Augmented Diffusion-Reaction (\textsc{IADR}). On the theoretical side, we establish a conditional $K$-step rollout error bound for this hypernetwork-conditioned operator, decomposing the long-horizon error into a one-step consistency term and a slice-wise stability term, and identifying the regimes under which the rollout error stays uniformly bounded.

We empirically validate the $\varepsilon$-generalisation of \textsc{IADR} on a suite of in-distribution and out-of-distribution evaluations of the Allen--Cahn equation. The in-distribution evaluation tests \textsc{IADR} on the same circular geometry and the same five training $\varepsilon$ values used at training time. The out-of-distribution evaluations test the model on unseen interface widths (both interpolating within the training interval $E$ and extrapolating beyond it), on unseen non-circular geometries, and on the combination of the two. A single \textsc{IADR} matches the per-$\varepsilon$ specialists on the in-distribution evaluation within the prescribed range, retains their cross-shape generalisation across all out-of-distribution shifts, and retains their compact diffusion-and-reaction (DR) target structure with millisecond-scale inference while remaining substantially faster than every general operator target.

In summary, our main contributions are (i) \textsc{IADR}, a neural operator for the Allen--Cahn equation: a target network with a diffusion-and-reaction structure, whose weights are produced by a hypernetwork from the interface-width parameter $\varepsilon$; (ii) a conditional $K$-step rollout error bound for this hypernetwork-conditioned operator, identifying a strict-contraction regime in which the long-horizon error stays uniformly bounded and an error-growth regime with an explicit bound; (iii) empirical validation showing that a single \textsc{IADR} matches the per-$\varepsilon$ specialists on in-distribution data within the prescribed range, retains their cross-shape generalisation on out-of-distribution geometries and unseen interface widths, and retains their compact DR target structure with millisecond-scale inference while remaining substantially faster than every general operator target.

\section{Related Work}
\label{sec:related}

\paragraph{Numerical schemes for mean-curvature flow.}
The Allen--Cahn equation has been studied as the phase-field model of mean-curvature flow for decades, and a mature line of semi-implicit time-stepping schemes is available. Convex splitting~\citep{eyre1998convex}, the semi-implicit Fourier-spectral method~\citep{chen1998applications}, and the diffusion and reaction (DR) splitting~\citep{bretin2022jcp} (the latter falling within the Lie--Trotter / Strang splitting family for gradient-flow PDEs~\citep{mclachlan2002splitting}) attain sub-grid accuracy with well understood stability and convergence properties; the systematic numerical analysis of mean-curvature flow and related geometric PDEs is surveyed in~\citep{deckelnick2005computation}, with the underlying sharp-interface theory grounded in the level-set formulation~\citep{evans1991motion} and the Allen--Cahn convergence results~\citep{ilmanen1993convergence}. The discretisations of these schemes, however, depend on $\varepsilon$ explicitly and must be re-instantiated for every interface width of interest, and the schemes are not differentiable end-to-end. These two limitations motivate the neural-network operators discussed next, which restore differentiability; the per-$\varepsilon$ limitation, however, persists in the neural-operator literature and motivates the present paper.

\paragraph{Operator learning.}
A second body of work learns the solution map of a PDE by a parametric neural network, within the broader physics-informed machine-learning paradigm~\citep{karniadakis2021physics, cuomo2022scientific}. Two threads are most relevant for the Allen--Cahn equation. \emph{General operator-learning frameworks}, consolidated by~\citet{kovachki2023neural}, learn solution maps that are not specific to any single PDE: families such as the Fourier Neural Operator~\citep{li2021fno}, DeepONet~\citep{lu2021deeponet}, and their graph and multipole variants~\citep{li2020gno,li2020mgno} cast the solution map as a function-to-function network with the PDE parameter concatenated to the input field, and physics-informed extensions such as PINO~\citep{li2021pino} layer a residual loss on top of this input-level conditioning, echoing the PINN framework~\citep{raissi2019pinn}. Applied to the Allen--Cahn equation, this thread covers the range of $\varepsilon$ within a single network, but the target carries no phase-field structure, so accuracy on physics-aligned metrics typically lags the operator-splitting schemes of the previous paragraph by a wide margin. \emph{PDE-specific operator learning} targets the Allen--Cahn equation directly: a common design is a black-box time-stepper that maps $u_t$ to $u_{t+1}$ on a fixed grid, often with a convolutional or U-Net backbone~\citep{wight2021solving}; physics penalties on a level-set or phase-field representation~\citep{mattey2022novel} and learned corrections to a numerical scheme~\citep{kochkov2021machine} extend this thread; the trainable DR splitting network of~\citet{bretin2022jcp} preserves the physical structure of the DR splitting but is bound to a single $\varepsilon$. Neither thread therefore delivers both $\varepsilon$-coverage and the phase-field structure of the target at the same time.

\paragraph{Meta-learning and hypernetworks for PDE solvers.}
A third line conditions a neural operator on a task descriptor through meta-learning or hypernetworks, with the aim of sharing weights across a family of PDE tasks. Meta-learning frameworks such as MAML~\citep{finn2017maml} and HyperNetworks~\citep{ha2017hypernetworks} have been adapted to PDE solvers in Meta-MgNet~\citep{chen2022metamgnet} and Hyper-LR-PINN~\citep{cho2023hyperlrpinn}, and subsequent work extends the line to foundation-scale PDE pre-training and task-adaptive conditioning~\citep{herde2024poseidon, hao2024dpot, koupai2024geps, bischof2025hypino, cao2024vicon}. The most direct realisation for our setting is DISCO~\citep{morel2025disco}: a space-time Transformer encodes a $T$-frame trajectory, a hypernetwork maps the resulting code to a full weight vector, and the weights are injected into a frozen U-Net target. None of these methods is specialised to the Allen--Cahn equation: the target network is designed to transfer across heterogeneous PDE families and therefore carries no phase-field structure, so their accuracy relative to the operator-learning schemes of the previous paragraph on the Allen--Cahn equation has not been established. Among the methods above, only DISCO yields a one-step neural propagator that can be rolled out under the same autoregressive protocol as \textsc{IADR}; Meta-MgNet conditions a multigrid smoother for linear elliptic systems and Hyper-LR-PINN fits a coordinate MLP $u(x,t)$ per initial condition, neither yielding an operator over $\mathcal{X}$. We therefore take DISCO as our head-to-head meta-learning baseline. The present paper fills this gap by specialising the target network to the Allen--Cahn equation's DR splitting structure, retaining phase-field fidelity while sharing parameters across $\varepsilon$.

\section{Method}
\label{sec:method}

\subsection{Problem Formulation}
\label{sec:method:problem}

We consider the Allen--Cahn equation as the phase-field model of mean-curvature flow on a domain $\Omega_x \subset \mathbb{R}^2$, where the order parameter $u : \Omega_x \times [0, T] \to \mathbb{R}$ evolves according to
\begin{equation}
\label{eq:allen-cahn}
\partial_t u = \Delta u - \tfrac{1}{\varepsilon^2} W'(u),
\qquad W(s) = \tfrac{1}{2} s^2 (1-s)^2 .
\end{equation}
For an evolving sharp interface $\partial \Omega(t)$ with signed distance function $d(\cdot, \Omega(t))$, the solution $u$ of \eqref{eq:allen-cahn} is known to concentrate in a layer of width $\varepsilon$ near $\partial\Omega(t)$; its theoretical idealisation is the exact phase-field representation
\begin{equation}
\label{eq:q-representation}
\mathbf{q}_\varepsilon(\cdot, t)
\;:=\; q\!\left( \tfrac{ d(\cdot, \Omega(t)) }{ \varepsilon } \right),
\qquad
q(r) = \tfrac{1}{2}(1 - \tanh(r/2)).
\end{equation}
This representation agrees with the solution $u$ of \eqref{eq:allen-cahn} up to a Hausdorff error of order $O(\varepsilon^{2}|\!\log\varepsilon|^{2})$ on the diffuse interface (Appendix~\ref{app:phase-field-setup}); we therefore take $\mathbf{q}_\varepsilon$ as the learning target. Its exact $\delta t$-propagator is denoted by
$\mathcal{T}_{\delta t,\varepsilon}(\mathbf{q}_\varepsilon(\cdot,t)) =
\mathbf{q}_\varepsilon(\cdot, t+\delta t)$. As the structural template of
our network we take the first-order DR splitting
\begin{equation}
\label{eq:dr-splitting-closed}
\mathcal{S}_{\delta t,\varepsilon}(\mathbf{q})
\;=\; \rho_{1,\varepsilon}(K_1 \ast \mathbf{q}),
\quad
\rho_{1,\varepsilon}(s) = s - \tfrac{\delta t}{\varepsilon^2} W'(s),
\quad
\widehat{K}_1(\xi) = \frac{1}{1 + \delta t\, 4\pi^2|\xi|^2}.
\end{equation}
The target network $\Phi$ of Section~\ref{sec:method:approach} mirrors $\mathcal{S}_{\delta t,\varepsilon}$ in topology: its weights are produced by the hypernetwork rather than derived analytically.
The $\Gamma$-convergence background and the $O(\delta t^{2})$ equivalence of the diffusion-then-reaction and reaction-then-diffusion orders are deferred to Appendices~\ref{app:phase-field-setup} and~\ref{app:proofs:order-equiv}.

Fix a compact computational domain $\Omega_x \subset \mathbb{R}^2$, time
step $\delta t > 0$, and prescribed interface-width range
$E = [\varepsilon_{\min}, \varepsilon_{\max}]$. Let $\mathcal{G}$ be a
class of admissible initial sets and let
$\mathcal{A}_\varepsilon = \{ \mathbf{q}_\varepsilon(\cdot, t) :
\Omega(0) \in \mathcal{G},\ t \in [0,T] \} \subset \mathcal{X}$,
$\mathcal{A} = \{ (\varepsilon, \mathbf{q}) : \varepsilon \in E,\
\mathbf{q} \in \mathcal{A}_\varepsilon \}$, with $\mathcal{X}$ the
$L^{2}(\Omega_x)$ phase-field space. We seek a single parameter vector
$\theta = \vartheta \in \mathbb{R}^m$ such that the hypernetwork-form
operator
\begin{equation}
\label{eq:dria-op}
\mathcal{P}^{\mathrm{NN}}_{\theta;\delta t}(\varepsilon, \mathbf{q})
\;=\; \Phi\big(\mathbf{q};\ h_\vartheta(\varepsilon)\big)
\end{equation}
approximates $\mathcal{T}_{\delta t,\varepsilon}$ on $\mathcal{A}$, with
$\Phi$ the target network of Section~\ref{sec:method:approach},
$h_\vartheta:\mathbb{R}\to\mathbb{R}^p$ a compact MLP reading the scalar
$\varepsilon$, and the slice-wise abbreviation
$\mathcal{N}_{\theta,\delta t,\varepsilon}(\mathbf{q}) :=
\Phi(\mathbf{q}; h_\vartheta(\varepsilon))$ used in
Section~\ref{sec:method:theory}. With $\mu_E$ on $E$ and
$\mu_\varepsilon$ on $\mathcal{A}_\varepsilon$, training minimises the
population one-step loss
\begin{equation}
\label{eq:population-loss}
\mathcal{L}(\theta)
\;=\; \int_E \int_{\mathcal{A}_\varepsilon}
\big\| \Phi(\mathbf{q}; h_\vartheta(\varepsilon))
- \mathcal{T}_{\delta t,\varepsilon}(\mathbf{q}) \big\|_{L^2(\Omega_x)}^2
\,\mathrm{d}\mu_\varepsilon(\mathbf{q})\,\mathrm{d}\mu_E(\varepsilon).
\end{equation}
The empirical objective, the Riemann-sum discretisation on the
$H \times W$ grid, the one-step rollout training, and the operator-level setup that motivates
the rollout analysis of Section~\ref{sec:method:theory} are collected in
Appendix~\ref{app:impl-details}.

\subsection{Proposed Approach}
\label{sec:method:approach}

\begin{figure}[t]
\centering
\includegraphics[width=0.95\linewidth]{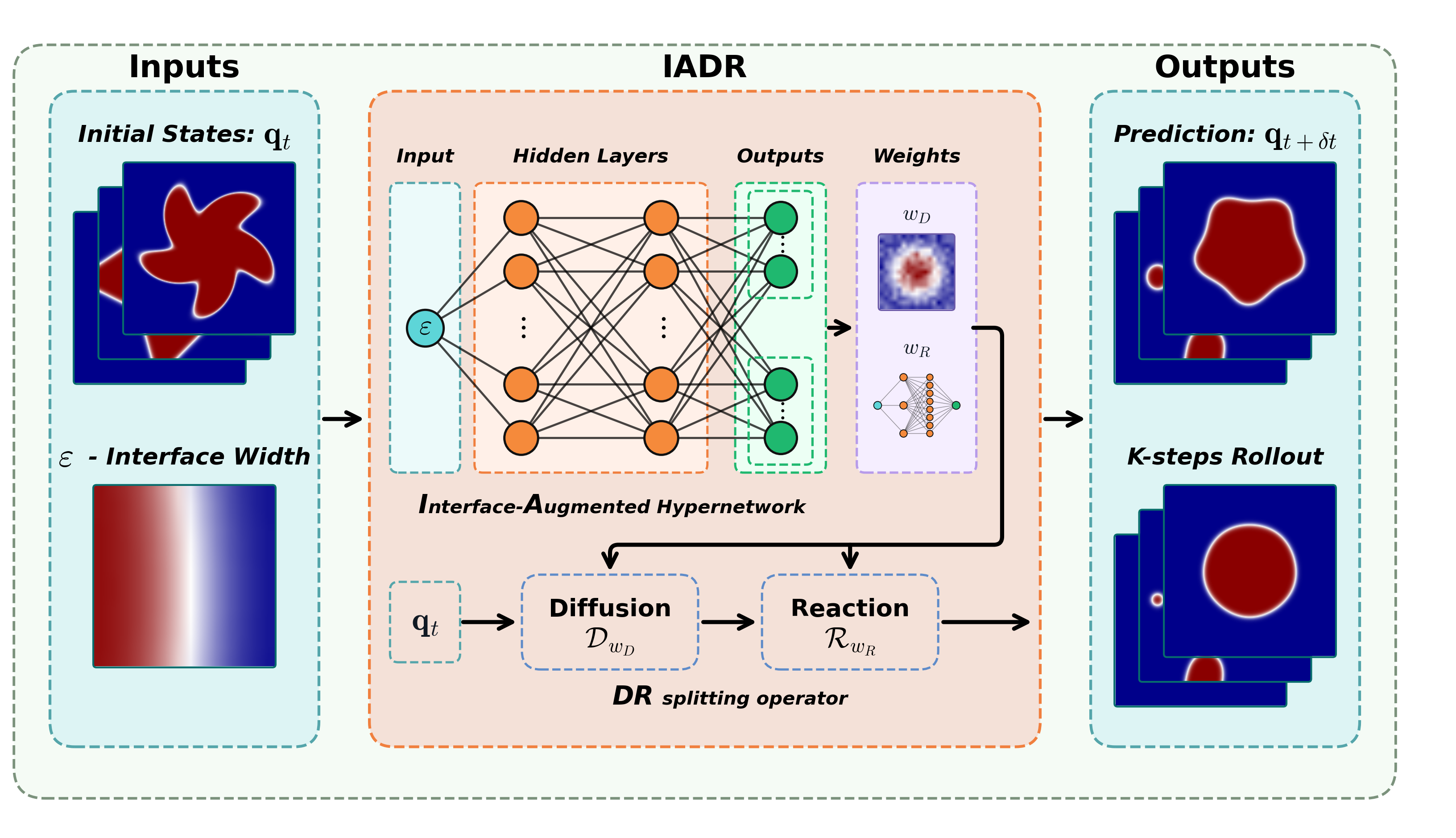}
\caption{Overview of \textsc{IADR}: the hypernetwork
$h_\vartheta(\varepsilon)$ emits the weights of the DR splitting
network $\Phi = \mathcal{R}_{w_R} \circ \mathcal{D}_{w_D}$ (target network), which
advances the phase field $\mathbf{q}_\varepsilon(\cdot, t)$ by one
time step and can be applied autoregressively for $K$-step rollout.}
\label{fig:iadr-overview}
\end{figure}

\paragraph{Approach overview.}
Inspired by the trainable DR splitting network of~\citet{bretin2022jcp} and by DISCO's hypernetwork-conditioning design~\citep{morel2025disco}, we construct \textsc{IADR} by coupling an MLP hypernetwork $h_\vartheta$ with a DR target network $\Phi$ that serves as the operator template. The hypernetwork generates the target network's weights from the interface width $\varepsilon$, and the target network uses these weights to evolve a given configuration. This design addresses the two requirements identified in Section~\ref{sec:related}: retaining the phase-field structure encoded in the target network's architecture while covering the range of $\varepsilon$ with a single model.

This separation is also empirically motivated: independently training $\Phi$ at the five training widths produces structured variation in its diffusion and reaction weights across $\varepsilon$ (Appendix~\ref{app:impl-details}, Figure~\ref{fig:weight-drift}), motivating our choice to generate both sets of weights from $\varepsilon$ through the hypernetwork.

\paragraph{Target network.}
The target network $\Phi(\cdot; w) : \mathcal{X} \to \mathcal{X}$
is the composition of a spatial diffusion block $\mathcal{D}_{w_D}$
and a pointwise reaction block $\mathcal{R}_{w_R}$, mirroring
the closed-form update
$\mathbf{q} \mapsto \rho_{1,\varepsilon}(K_1 \ast \mathbf{q})$
of~\eqref{eq:dr-splitting-closed}:
\begin{equation}
\label{eq:dria-phi-decomp}
\Phi(\mathbf{q}; w)
\;=\; \mathcal{R}_{w_R}\!\left( \mathcal{D}_{w_D}(\mathbf{q}) \right),
\qquad
w = (w_R, w_D) \in \mathbb{R}^{p_R + p_D} =: \mathbb{R}^p.
\end{equation}
Here, $w_D$ and $w_R$ are the weight vectors of the spatial diffusion and pointwise reaction blocks, respectively, both generated by the hypernetwork $h_\vartheta$ from the interface width $\varepsilon$, and their dimensions are $p_D$ and $p_R$, respectively.
For every $\varepsilon$, the operator is $\Phi(\cdot; h_\vartheta(\varepsilon))$, and the whole family is parameterised by the hypernetwork parameter vector $\vartheta$, which contains all trainable parameters, together with the fixed structure of the target network $\Phi$. No explicit $\varepsilon$ channel
enters $\Phi$ itself; the $\varepsilon$-dependence is concentrated in
$h_\vartheta$.
We next describe the two blocks of the DR operator and the hypernetwork in more detail.

The diffusion block $\mathcal{D}_{w_D}$ is a single $K_D \times K_D$
two-dimensional convolution acting on $\mathbf{q}$ in the spatial
domain, with kernel weights $w_D \in \mathbb{R}^{K_D^{2}}$ emitted by
$h_\vartheta$. For a given $\varepsilon$, the same kernel acts at every grid point, coupling spatial locations through a linear weighted sum of neighbouring phase-field values. Its output is an intermediate field on the same grid, which is then passed to the pointwise reaction block. Structurally it is the discrete analogue of the
convolution $K_1 \ast \mathbf{q}$ in the closed-form first-order DR
step~\eqref{eq:dr-splitting-closed}, where $K_1$ is the spatial kernel
of the Helmholtz resolvent $(I - \delta t \Delta)^{-1}$. The block has
no bias and no spatial-channel mixing; its capacity is exactly the
$K_D^{2}$ kernel entries, all generated by the hypernetwork. The
concrete kernel size $K_D$ is collected in
Appendix~\ref{app:impl-details}.

The reaction block $\mathcal{R}_{w_R}$ is a compact scalar-to-scalar
multilayer perceptron $\mathrm{MLP} : \mathbb{R} \to \mathbb{R}$
acting pointwise on $\mathbf{q}$, and for every $x \in \Omega_x$,
$(\mathcal{R}_{w_R} \mathbf{q})(x) = \mathrm{MLP}(\mathbf{q}(x))$.
Within the target network $\Phi$, this block receives the intermediate field produced by the diffusion block. At each grid point, the hidden layers transform the scalar phase-field input into nonlinear features, which the output layer maps back to a scalar phase-field value. For a given $\varepsilon$, the same MLP is shared across all grid points, so this step updates each value separately without mixing neighbouring values.
Its weight vector
$w_R \in \mathbb{R}^{p_R}$ is emitted by $h_\vartheta$. We choose an MLP as
the learned analogue of the cubic reaction
$\rho_{1,\varepsilon}(s) = s - (\delta t / \varepsilon^{2}) W'(s)$
because hard-coding the cubic form would commit the network to the analytic
profile of $W'$, whereas a small MLP allows the reaction profile to
adapt to $\varepsilon$ and to absorb mild discretisation artefacts at
the working amplitude band, while the scalar-to-scalar locality and
translation invariance of the classical reaction step are preserved by
construction.

\paragraph{Hypernetwork.}
The hypernetwork $h_\vartheta : \mathbb{R} \to \mathbb{R}^{p_R + p_D}$
is a multilayer perceptron with $L_h$ hidden layers, hidden width $h$,
smooth activation, scalar input $\varepsilon$ and output dimension
equal to the full weight count of $\Phi$:
\begin{equation}
\label{eq:dria-hypernet}
h_\vartheta(\varepsilon) \;=\; \alpha \cdot \mathrm{MLP}_\vartheta(\varepsilon),
\end{equation}
where $\alpha > 0$ is a fixed scalar multiplier applied to the raw
MLP output to keep the emitted weights at a workable scale; its
concrete value and the network shapes are collected in
Appendix~\ref{app:impl-details}. The trainable parameters
of \textsc{IADR} are exactly $\vartheta$; the target network
$\Phi$ has no independently trainable parameters, and at every forward pass its weights are
set to $h_\vartheta(\varepsilon)$ via the weight-injection mechanism of DISCO, after which gradients flow through both
networks. Capacity is therefore concentrated in $\vartheta$, while the
target network $\Phi$ remains the same compact two-block stack at
every $\varepsilon$. The same hypernetwork interface
extends to higher-order DR splittings; details are deferred to
Appendix~\ref{app:higher-order}.

\subsection{Theoretical Analysis}
\label{sec:method:theory}

Neural operators such as \textsc{IADR} advance the phase field one time step at a time; accurate long-horizon prediction therefore requires that single-step errors remain controlled over autoregressive rollouts of many steps. We establish a conditional $K$-step rollout error bound in terms of uniform one-step consistency $\tau$ and slice-wise Lipschitz stability $L$. The bound identifies a strict-contraction regime ($L < 1$), in which the rollout error stays uniformly bounded in $K$, and an error-growth regime ($L \geq 1$), under the stated assumptions.

\begin{theorem}[\texorpdfstring{$K$}{K}-step rollout stability, informal]
\label{thm:rollout-main}
For each $\varepsilon \in E$, let
$\mathcal{T}_{\delta t,\varepsilon},\,
 \mathcal{N}_{\theta,\delta t,\varepsilon} :
 \mathcal{A}_\varepsilon \to \mathcal{A}_\varepsilon$
be the exact and learned slice-wise propagators. Assume there exist
constants $L_T \geq 0$, $L_N \geq 0$, and $\tau \geq 0$, independent of
$\varepsilon$, such that for every $\varepsilon \in E$ and every
$\mathbf{q}, \mathbf{q}' \in \mathcal{A}_\varepsilon$,
\[
\big\| \mathcal{T}_{\delta t,\varepsilon}(\mathbf{q})
      - \mathcal{T}_{\delta t,\varepsilon}(\mathbf{q}') \big\|_{\mathcal{X}}
\leq
L_T \, \|\mathbf{q} - \mathbf{q}'\|_{\mathcal{X}},
\]
\[
\big\| \mathcal{N}_{\theta,\delta t,\varepsilon}(\mathbf{q})
      - \mathcal{N}_{\theta,\delta t,\varepsilon}(\mathbf{q}') \big\|_{\mathcal{X}}
\leq
L_N \, \|\mathbf{q} - \mathbf{q}'\|_{\mathcal{X}},
\]
\[
\big\| \mathcal{N}_{\theta,\delta t,\varepsilon}(\mathbf{q})
      - \mathcal{T}_{\delta t,\varepsilon}(\mathbf{q}) \big\|_{\mathcal{X}}
\leq \tau.
\]
Set $L := \max\{L_T, L_N\}$. Then for every integer $K \geq 1$,
\[
\sup_{\varepsilon \in E}
\sup_{\mathbf{q} \in \mathcal{A}_\varepsilon}
\big\| \mathcal{T}_{\delta t,\varepsilon}^K(\mathbf{q})
      - \mathcal{N}_{\theta,\delta t,\varepsilon}^K(\mathbf{q}) \big\|_{\mathcal{X}}
\;\leq\;
\tau \sum_{j=0}^{K-1} L^j.
\]
\end{theorem}

\begin{corollary}[Two regimes]
\label{cor:rollout-regimes}
Under the assumptions of Theorem~\ref{thm:rollout-main}, the bound
$\tau \sum_{j=0}^{K-1} L^{j}$ specialises in two regimes:
\begin{itemize}
\item[(i)] if $L \geq 1$, the bound reduces to $\tau K\, L^{K-1}$, which
 grows linearly in $K$ when $L = 1$ and at geometric rate $L^{K-1}$
 when $L > 1$;
\item[(ii)] if $L < 1$, the geometric series converges and
 \[
 \sup_{\varepsilon \in E}
 \sup_{\mathbf{q} \in \mathcal{A}_\varepsilon}
 \big\| \mathcal{T}_{\delta t,\varepsilon}^K(\mathbf{q})
       - \mathcal{N}_{\theta,\delta t,\varepsilon}^K(\mathbf{q}) \big\|_{\mathcal{X}}
 \;\leq\; \frac{\tau}{1 - L},
 \]
 uniformly in $K$, with residual decaying to zero at geometric rate
 $L^{K}$.
\end{itemize}
\end{corollary}

\paragraph{Proof sketch.} The argument is a discrete error recursion
on each slice $\mathcal{A}_\varepsilon$.

\emph{Step 1 (one-step error).} Under
Assumption~\ref{assum:consistency}, the one-step prediction error
$\|\mathcal{T}_{\delta t,\varepsilon}(q) -
  \mathcal{N}_{\theta,\delta t,\varepsilon}(q)\|_{\mathcal{X}}
  \leq \tau$ holds uniformly for $q \in \mathcal{A}_\varepsilon$.

\emph{Step 2 (slice-wise stability).} Under
Assumptions~\ref{assum:lip-T}--\ref{assum:lip-N}, both
$\mathcal{T}_{\delta t,\varepsilon}$ and
$\mathcal{N}_{\theta,\delta t,\varepsilon}$ are $L$-Lipschitz on
$\mathcal{A}_\varepsilon$ with $L = \max\{L_T, L_N\}$, so the
accumulated error is amplified by at most $L$ per step.

\emph{Step 3 (telescoping).} Iterating Steps 1--2 yields
$e_K \leq \tau \sum_{k=0}^{K-1} L^{k}$, which gives the strict-contraction regime ($L<1$, error $\leq \tau / (1-L)$ uniform in $K$)
and the error-growth regime ($L \geq 1$, error $\leq \tau\, K\, L^{K-1}$)
of Corollary~\ref{cor:rollout-regimes}. The resulting estimate
explains why long-horizon rollout is governed jointly by one-step
consistency $\tau$ and by the slice-wise stability $L$ of the learned
operator family. The formal version with problem-level assumptions
is given in Appendix~\ref{app:proofs:slice}; smaller slice-wise
Lipschitz constants $L_N$ tighten the bound.

\paragraph{Theory and experiments.}
Under the stated assumptions, the bound relates one-step consistency and slice-wise stability to error accumulation, guiding the design of autoregressive rollout evaluation using worst-frame $L^{2}$-$T^{\infty}$ errors. Tests at unseen interface widths and on non-circular geometries assess long-horizon accuracy under parameter and geometry shifts. The lemma in Appendix~\ref{app:proofs:order-equiv} establishes first-order consistency of both DR splitting orders, supporting DR splitting as the target-network template.

\section{Experiments}
\label{sec:exp}

\subsection{Experimental Setup}
\label{sec:exp:setup}

\paragraph{Setup and tasks.}
We evaluate \textsc{IADR} on the Allen--Cahn equation on the flat torus
$\Omega_x = [0,1]^2$, assessing generalisation along two axes:
interface width and geometry. This gives one in-distribution evaluation
and three out-of-distribution evaluations:
\emph{Task~1} circles at $E_{\text{train}}$ (in-distribution);
\emph{Task~2} circles at $\{0.0069, 0.0081, 0.0093\}$ ($\varepsilon$-OOD);
\emph{Task~3} non-circular geometries at $\varepsilon = 0.0078 \in E_{\text{train}}$
(geometry-OOD); and
\emph{Task~4} non-circular geometries at $\{0.0069, 0.0093\}$ (joint OOD).
The domain is discretised on a $256 \times 256$ periodic grid with
time step $\delta t = 6.10 \times 10^{-5}$, chosen to remain in the
stability region of the explicit DR splitting across the interface-width range $E = [0.0065, 0.0090]$.
The $\varepsilon$-axis is sampled at the five training widths
$E_{\text{train}} = \{0.0065, 0.0072, 0.0078, 0.0084, 0.0090\} \subset E$,
two in-distribution interpolation widths $\{0.0069, 0.0081\} \subset E$,
and one OOD-coarse width $\varepsilon = 0.0093 \notin E$
($\varepsilon > \varepsilon_{\max}$, outside the training interval but
still an $\varepsilon$-OOD evaluation).
Training uses only circular initial sets at the five $E_{\text{train}}$ widths;
no non-circular geometry and no width outside $E_{\text{train}}$ is seen
at training time.
Circular trajectories follow the closed-form law $R(t) = \sqrt{R_0^2 - 2t}$,
used as the exact reference for both training and evaluation.
The non-circular geometries are \emph{spiral disks}, disks deformed by a
small radial ripple along the polar angle, and \emph{five-pointed stars},
a central disk with five evenly spaced triangular arms; both are used only
at evaluation time, with reference trajectories computed by the higher-order
DR splitting scheme.
Each model takes a single conditioning frame, except DISCO, which uses $16$ consecutive ground-truth frames, and rolls out autoregressively
to a fixed horizon $T_\text{eval}$. Initial-set parameters, sample counts,
and trajectory budgets for all three datasets are listed in
Appendix~\ref{app:exp-details:datasets}, Table~\ref{tab:dataset-config}.

\paragraph{Baselines.}
We select baselines from the three lines reviewed in
Section~\ref{sec:related}. \emph{For PDE-specific operator learning},
we use \emph{Per-$\varepsilon$ DR}: an independent DR splitting
network~\citep{bretin2022jcp} retrained at each of the five training
widths, serving as our expert (SOTA) reference. \emph{For general
operator learning}, we use \emph{FNO}~\citep{li2021fno}, a Fourier
neural operator trained on the same data with $\varepsilon$
concatenated to the input channel. \emph{For meta-learning over PDE
operators}, we use \emph{DISCO}~\citep{morel2025disco}: the original
DISCO pipeline unchanged, in which a space-time Transformer encodes
a $T$-frame trajectory and a hypernetwork maps the resulting code
into a frozen U-Net target. We additionally include an in-house
\emph{U-Net variant} that swaps \textsc{IADR}'s diffusion-and-reaction
target for the four-level U-Net of~\citep{morel2025disco} while
keeping the rest of the conditioning pipeline; this isolates the
role of \textsc{IADR}'s physics-structured target by holding the
conditioning pipeline fixed and is not a representative of any
external category. DISCO trains for $100$ epochs and all other models for $20{,}000$ epochs.
Architectures and full hyperparameters are listed in
Appendix~\ref{app:impl-details}. Two further ablations on the role
of $\varepsilon$ and on the conditioning mechanism are reported in
Appendix~\ref{app:ablation-results}.

\paragraph{Metrics.}
To assess whether a learned operator is physically faithful over an
autoregressive rollout, we report three classes of metric on each
trajectory.
\textbf{(i) $L^{2}$ accuracy.} The $L^{2}$-$T^{\infty}$ error,
taken from~\citet{bretin2022jcp}, is the worst-frame physical $L^{2}$
error per trajectory; it
quantifies how far the predicted phase field strays from the
reference solution at the worst moment of the rollout.
\textbf{(ii) Geometric mass conservation.} The Volume-$T^{\infty}$
error, also taken from~\citet{bretin2022jcp}, is the worst-frame
deviation of the predicted enclosed mass from that of the reference
solution; on the circle dataset it coincides with the analytical
mean-curvature decay $A'(t) = -2\pi$, so it directly tests whether
the model respects the geometric mass-loss law of the underlying
mean-curvature flow.
\textbf{(iii) Diffuse-interface consistency.} The AIW metric, which
we introduce, is the deviation of
the predicted average interface width from its
$\varepsilon$-controlled target at a fixed late-time frame; it
certifies that the diffuse-interface width remains consistent
with the prescribed $\varepsilon$ rather than spreading or
sharpening unphysically. We add AIW because the existing $L^{2}$
and volume metrics do not directly penalise drift in the diffuse
layer.
Discrete formulas and aggregation rules are given in
Appendix~\ref{app:exp-details:metrics}.

\subsection{Main Results}
\label{sec:exp:results}

Table~\ref{tab:main} reports the three physics-aligned metrics on the
four-task suite, and Table~\ref{tab:step-time} the single-step
inference cost; full $\mathrm{mean} \pm \mathrm{std}$ in
Table~\ref{tab:main-full}.

\begin{table}[t]
\centering
\scriptsize
\setlength{\tabcolsep}{2pt}
\caption{Main results. Best result per row in bold, second best underlined.}
\label{tab:main}
\resizebox{\linewidth}{!}{%
\begin{tabular}{ll*{15}{c}}
\toprule
 & & \multicolumn{5}{c}{$L^{2}\text{-}T^{\infty}$ ($\times 10^{-3}$)}
 & \multicolumn{5}{c}{$V\text{-}T^{\infty}$ ($\times 10^{-3}$)}
 & \multicolumn{5}{c}{AIW ($\times 10^{-3}$)} \\
\cmidrule(lr){3-7} \cmidrule(lr){8-12} \cmidrule(lr){13-17}
Shape & $\varepsilon$
 & \textsc{IADR} & per-$\varepsilon$ DR & U-Net & FNO & DISCO
 & \textsc{IADR} & per-$\varepsilon$ DR & U-Net & FNO & DISCO
 & \textsc{IADR} & per-$\varepsilon$ DR & U-Net & FNO & DISCO \\
\midrule
\multicolumn{17}{c}{\textit{Task 1: circles at training widths (in-distribution)}} \\
\midrule
 \multirow{5}{*}{Circle} & $0.0065$ & $\underline{33.98}$ & $\mathbf{33.75}$ & $921.67$ & $518.77$ & diverged & $\underline{3.81}$ & $\mathbf{3.79}$ & $680.39$ & $261.42$ & diverged & $\mathbf{51.84}$ & $\underline{51.87}$ & $69.46$ & $76.61$ & diverged \\
  & $0.0072$ & $\underline{27.06}$ & $\mathbf{18.89}$ & $951.33$ & $510.70$ & $6.7\!\times\!10^{13}$ & $\underline{2.97}$ & $\mathbf{2.09}$ & $709.87$ & $269.69$ & $7.2\!\times\!10^{12}$ & $\underline{0.34}$ & $\mathbf{0.32}$ & $25.62$ & $24.51$ & $398.57$ \\
  & $0.0078$ & $\underline{22.35}$ & $\mathbf{6.93}$ & $959.88$ & $508.08$ & $4.7\!\times\!10^{13}$ & $\underline{2.37}$ & $\mathbf{0.92}$ & $728.62$ & $269.20$ & $5.7\!\times\!10^{12}$ & $\underline{0.48}$ & $\mathbf{0.37}$ & $17.45$ & $20.81$ & $411.65$ \\
  & $0.0084$ & $\underline{14.78}$ & $\mathbf{4.53}$ & $974.31$ & $508.02$ & $2.5\!\times\!10^{12}$ & $\underline{1.39}$ & $\mathbf{0.94}$ & $745.59$ & $264.67$ & $2.3\!\times\!10^{11}$ & $\underline{0.54}$ & $\mathbf{0.35}$ & $23.25$ & $16.86$ & $419.19$ \\
  & $0.0090$ & $\underline{11.80}$ & $\mathbf{7.36}$ & $984.91$ & $509.65$ & $1.6\!\times\!10^{18}$ & $\underline{1.18}$ & $\mathbf{0.84}$ & $755.55$ & $268.61$ & $3.5\!\times\!10^{17}$ & $\underline{1.00}$ & $\mathbf{0.50}$ & $27.56$ & $14.84$ & $409.46$ \\
\midrule
\multicolumn{17}{c}{\textit{Task 2: circles at non-training widths ($\varepsilon$-OOD)}} \\
\midrule
 \multirow{3}{*}{Circle} & $0.0069$ & $\underline{28.08}$ & $\mathbf{25.46}$ & $937.71$ & $508.98$ & $1.1\!\times\!10^{10}$ & $\underline{3.08}$ & $\mathbf{2.82}$ & $698.39$ & $265.67$ & $8.7\!\times\!10^{8}$ & $\mathbf{0.37}$ & $\underline{0.41}$ & $24.17$ & $29.90$ & $396.72$ \\
  & $0.0081$ & $\underline{19.99}$ & $\mathbf{9.24}$ & $968.03$ & $500.88$ & $7.1\!\times\!10^{9}$ & $\underline{2.07}$ & $\mathbf{1.16}$ & $737.03$ & $265.36$ & $7.1\!\times\!10^{8}$ & $\underline{0.64}$ & $\mathbf{0.50}$ & $20.79$ & $20.59$ & $416.88$ \\
  & $0.0093$ & $\underline{65.95}$ & $\mathbf{8.27}$ & $778.98$ & $510.99$ & $3.1\!\times\!10^{11}$ & $\underline{9.97}$ & $\mathbf{0.95}$ & $549.11$ & $269.45$ & $4.8\!\times\!10^{10}$ & $16.76$ & $\mathbf{0.85}$ & $40.81$ & $\underline{12.11}$ & $410.05$ \\
\midrule
\multicolumn{17}{c}{\textit{Task 3: non-circular geometries at training widths (geometry-OOD)}} \\
\midrule
 \multirow{1}{*}{Five-pointed star} & $0.0078$ & $\underline{21.85}$ & $\mathbf{8.80}$ & $757.35$ & $553.48$ & $406.12$ & $\underline{3.77}$ & $\mathbf{1.62}$ & $603.77$ & $229.99$ & $224.92$ & $\underline{1.98}$ & $\mathbf{1.87}$ & $11.52$ & $19.34$ & $184.83$ \\
 \multirow{1}{*}{Spiral disk} & $0.0078$ & $\underline{4.03}$ & $\mathbf{3.65}$ & $119.04$ & $382.58$ & $205.19$ & $\underline{1.08}$ & $\mathbf{0.96}$ & $33.76$ & $86.77$ & $16.24$ & $0.44$ & $\underline{0.35}$ & $\mathbf{0.23}$ & $18.04$ & $295.82$ \\
\midrule
\multicolumn{17}{c}{\textit{Task 4: non-circular geometries at non-training widths (joint OOD)}} \\
\midrule
 \multirow{2}{*}{Five-pointed star} & $0.0069$ & $\underline{16.73}$ & $\mathbf{13.35}$ & $720.87$ & $556.31$ & $407.88$ & $\underline{2.81}$ & $\mathbf{2.25}$ & $547.74$ & $231.35$ & $222.74$ & $\mathbf{1.70}$ & $\underline{1.71}$ & $7.95$ & $20.80$ & $180.18$ \\
  & $0.0093$ & $\underline{50.26}$ & $\mathbf{28.34}$ & $790.93$ & $550.19$ & $393.26$ & $\underline{10.75}$ & $\mathbf{5.23}$ & $659.78$ & $229.31$ & $222.89$ & $\mathbf{2.33}$ & $\underline{2.38}$ & $9.06$ & $13.43$ & $186.33$ \\
 \multirow{2}{*}{Spiral disk} & $0.0069$ & $\underline{7.71}$ & $\mathbf{7.59}$ & $112.61$ & $373.37$ & $213.23$ & $\underline{1.97}$ & $\mathbf{1.95}$ & $27.20$ & $82.75$ & $16.06$ & $\underline{0.34}$ & $0.36$ & $\mathbf{0.19}$ & $24.03$ & $286.82$ \\
  & $0.0093$ & $\underline{14.16}$ & $\mathbf{3.72}$ & $170.10$ & $382.93$ & $205.85$ & $\underline{4.24}$ & $\mathbf{0.64}$ & $58.13$ & $88.09$ & $16.00$ & $0.56$ & $\underline{0.56}$ & $\mathbf{0.49}$ & $8.02$ & $293.18$ \\
\bottomrule
\end{tabular}%
}
\end{table}

\paragraph{\textsc{IADR} matches the per-$\varepsilon$ DR specialists
on accuracy and physical consistency within the prescribed range while every generic operator
target lags behind.}
On Task~1 (circles at training widths), \textsc{IADR}'s
$L^{2}\text{-}T^{\infty}$, $V\text{-}T^{\infty}$, and AIW closely
match the matching per-$\varepsilon$ DR specialist, with the
specialist attaining the smaller mean in five of five widths on
$L^{2}\text{-}T^{\infty}$ and $V\text{-}T^{\infty}$ and in four of
five widths on AIW. On Task~2 (circles at non-training widths in
$E$), \textsc{IADR} also recovers the specialist level on the two
interpolation widths $\{0.0069, 0.0081\}$ across every metric. By
contrast, every generic operator target sits well above the
per-$\varepsilon$ specialist on the same cells: at circle-$0.0078$,
\textsc{IADR} reads $L^{2}\text{-}T^{\infty} = 22.35 \times 10^{-3}$
and the specialist $6.93 \times 10^{-3}$, while FNO reads
$508.08 \times 10^{-3}$ and the U-Net target $959.88 \times 10^{-3}$;
$V$-$T^{\infty}$ and AIW follow the same pattern. For DISCO,
$L^{2}\text{-}T^{\infty}$ reaches $4.7 \times 10^{10}$. The DR splitting target therefore
gives the network the level of physical consistency only a
per-$\varepsilon$ specialist achieves, and the
$\varepsilon$-conditioning lets a single \textsc{IADR} hold this
level across the range, realising contribution~(iii) of
Section~\ref{sec:intro}. On the OOD-coarse width
$\varepsilon = 0.0093$, by contrast, \textsc{IADR}'s
$L^{2}\text{-}T^{\infty}$ jumps to $65.95 \times 10^{-3}$, well
above the per-$\varepsilon$ specialist's $8.27 \times 10^{-3}$. To further examine the need for full DR weight generation and evaluate \textsc{IADR}'s ability to cover a wider $\varepsilon$ range, we also compare \textsc{IADR} with Direct-$\varepsilon$-DR and specialist weight interpolation on circles, and train and evaluate \textsc{IADR} over a wider range (see Appendix~\ref{app:more-results}~\&~\ref{app:ablation-results}).

\paragraph{\textsc{IADR} preserves the per-$\varepsilon$ specialists'
cross-shape generalisation when transferred to non-circular
geometries.}
Tasks~3 and~4 transfer the trained model to non-circular geometries
(five-pointed star and spiral disk), neither seen during training.
On Task~3 (training widths), \textsc{IADR} stays close to the
matching-$\varepsilon$ specialist on every cell across all three
metrics. On Task~4 (non-training widths), \textsc{IADR} extends this
behaviour into the joint OOD regime: at the OOD-coarse width
$\varepsilon = 0.0093$, the per-$\varepsilon$ specialist outperforms
\textsc{IADR} on five-pointed star ($28.34 \times 10^{-3}$ vs
$50.26 \times 10^{-3}$ on $L^{2}\text{-}T^{\infty}$); on spiral disk
at the same width, \textsc{IADR} performs worse than the specialist
($14.16 \times 10^{-3}$ vs $3.72 \times 10^{-3}$) but remains well
below every generic operator target on this cell (U-Net
$170.10 \times 10^{-3}$, FNO $382.93 \times 10^{-3}$, DISCO
$205.85 \times 10^{-3}$). Across almost every Task~3 and Task~4 cell
the generic operator targets stay consistently above \textsc{IADR}
and the per-$\varepsilon$ specialists. A single $\varepsilon$-conditional
model therefore inherits the per-$\varepsilon$ specialist's level of
cross-shape generalisation rather than trading it for
$\varepsilon$-coverage, completing contribution~(iii).

\paragraph{\textsc{IADR} attains this accuracy at a notably lower
per-step inference cost than every generic operator target.}
Table~\ref{tab:step-time} reports the single-step forward time on a
$256 \times 256$ grid, aggregated by shape. \textsc{IADR} runs at
roughly $1.3$\,ms per step across all three shape families,
consistently faster than the U-Net target reproduced under the same
conditioning pipeline ($4.6$\,ms), the FNO general operator
($3.3$--$3.4$\,ms), and DISCO ($105$--$119$\,ms; about $80$--$90\times$
slower than \textsc{IADR}, driven by the space-time Transformer
encoder over the $T$-frame trajectory). \textsc{IADR} retains the compact DR target structure of the per-$\varepsilon$ specialists and covers the range with one model, avoiding retraining for new $\varepsilon$ values within that range. At inference, the additional weight generation and injection lead to a single-step time of $1.27$--$1.36$\,ms, compared with $0.61$--$0.62$\,ms for the specialists. Both remain at the millisecond scale and are substantially faster than the generic operator targets. Despite this additional per-step cost, training one \textsc{IADR} takes $16$\,h, compared with $24$\,h in total for the eight specialists in our comparison, demonstrating its lower total training cost across multiple $\varepsilon$ values.

\begin{table}[h]
\centering
\caption{Single-step inference time (ms/sample on $256\!\times\!256$, mean $\pm$ std across $\varepsilon$ within each shape). Best result per row in bold, second best underlined.}
\label{tab:step-time}
\begin{tabular}{lccccc}
\toprule
shape & \textsc{IADR} & per-$\varepsilon$ DR & U-Net & FNO & DISCO \\
\midrule
circle             & $\underline{1.30 \pm 0.02}$ & $\mathbf{0.61 \pm 0.02}$ & $4.63 \pm 0.09$ & $3.34 \pm 0.01$ & $119.13 \pm 3.55$ \\
five-pointed star  & $\underline{1.27 \pm 0.01}$ & $\mathbf{0.62 \pm 0.02}$ & $4.56 \pm 0.02$ & $3.38 \pm 0.01$ & $107.29 \pm 1.77$ \\
spiral disk        & $\underline{1.36 \pm 0.11}$ & $\mathbf{0.62 \pm 0.03}$ & $4.60 \pm 0.08$ & $3.38 \pm 0.00$ & $104.62 \pm 2.68$ \\
\bottomrule
\end{tabular}
\end{table}

\subsection{Case Study}
\label{sec:exp:case-study}

To complement the aggregate trends in Table~\ref{tab:main} we visualise
representative rollouts that anchor the findings of
Section~\ref{sec:exp:results} to single-frame qualitative behaviour.

\begin{figure}[t]
  \centering
  \begin{subfigure}[c]{0.54\linewidth}
    \begin{minipage}[t]{0.10\linewidth}\subcaption{}\label{fig:case-circle-mid}\end{minipage}%
    \raisebox{-\height}{\includegraphics[width=0.88\linewidth]{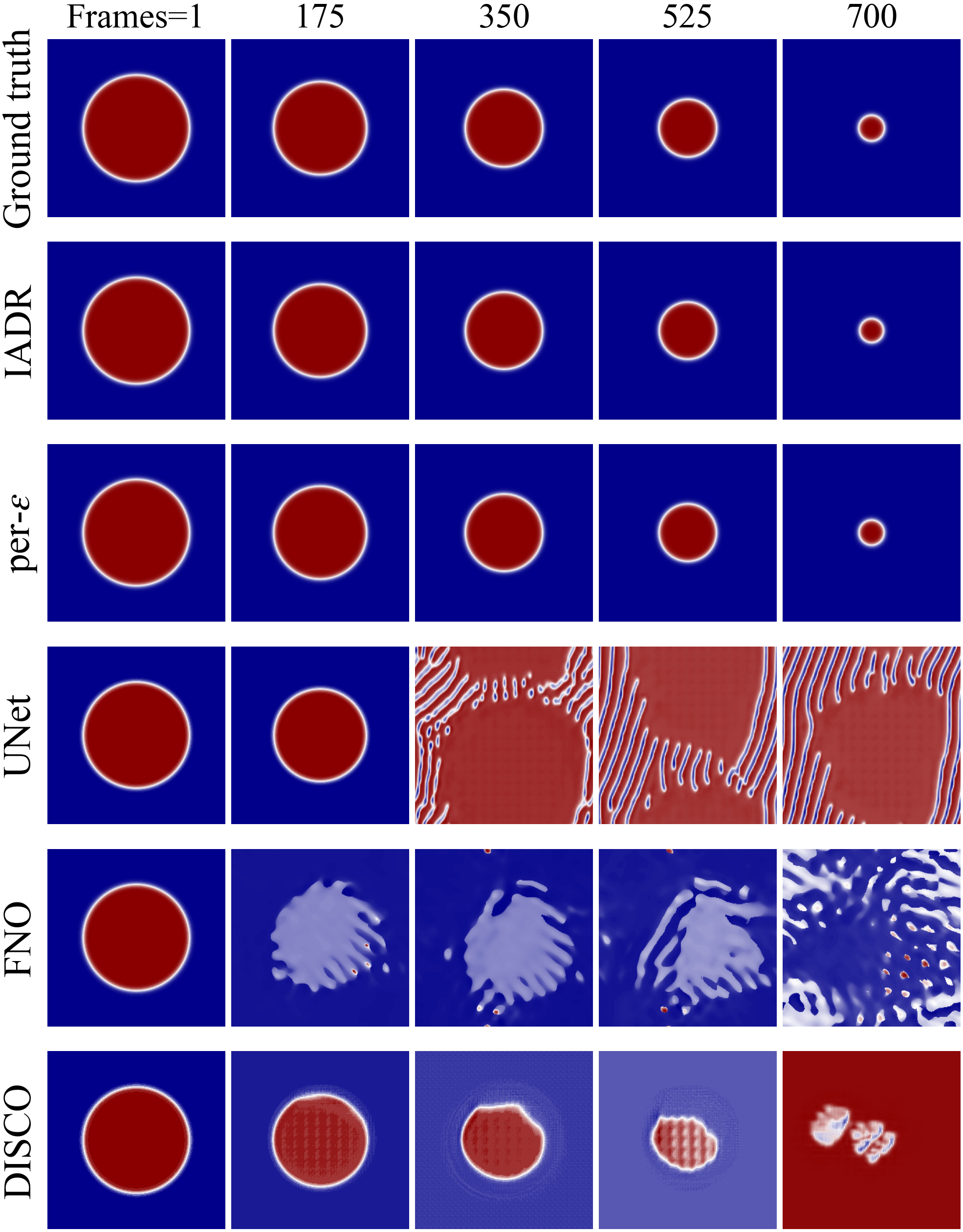}}%
  \end{subfigure}\hspace{4pt}%
  \begin{subfigure}[c]{0.44\linewidth}
    \begin{minipage}[t]{0.10\linewidth}\subcaption{}\label{fig:case-fs}\end{minipage}%
    \raisebox{-\height}{\includegraphics[width=0.88\linewidth]{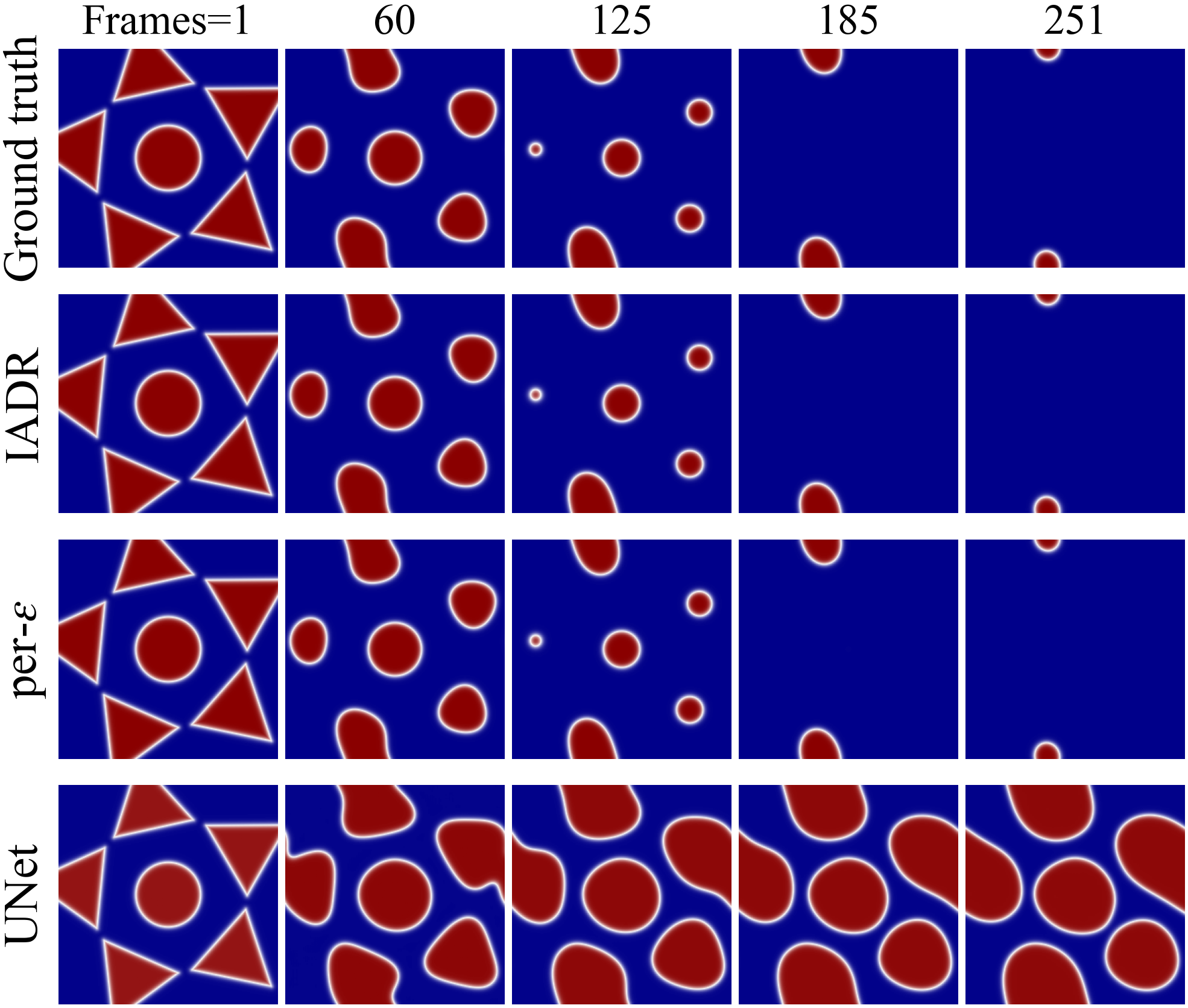}}%
    \par\vspace{6pt}
    \begin{minipage}[t]{0.10\linewidth}\subcaption{}\label{fig:case-circle-coarse}\end{minipage}%
    \raisebox{-\height}{\includegraphics[width=0.88\linewidth]{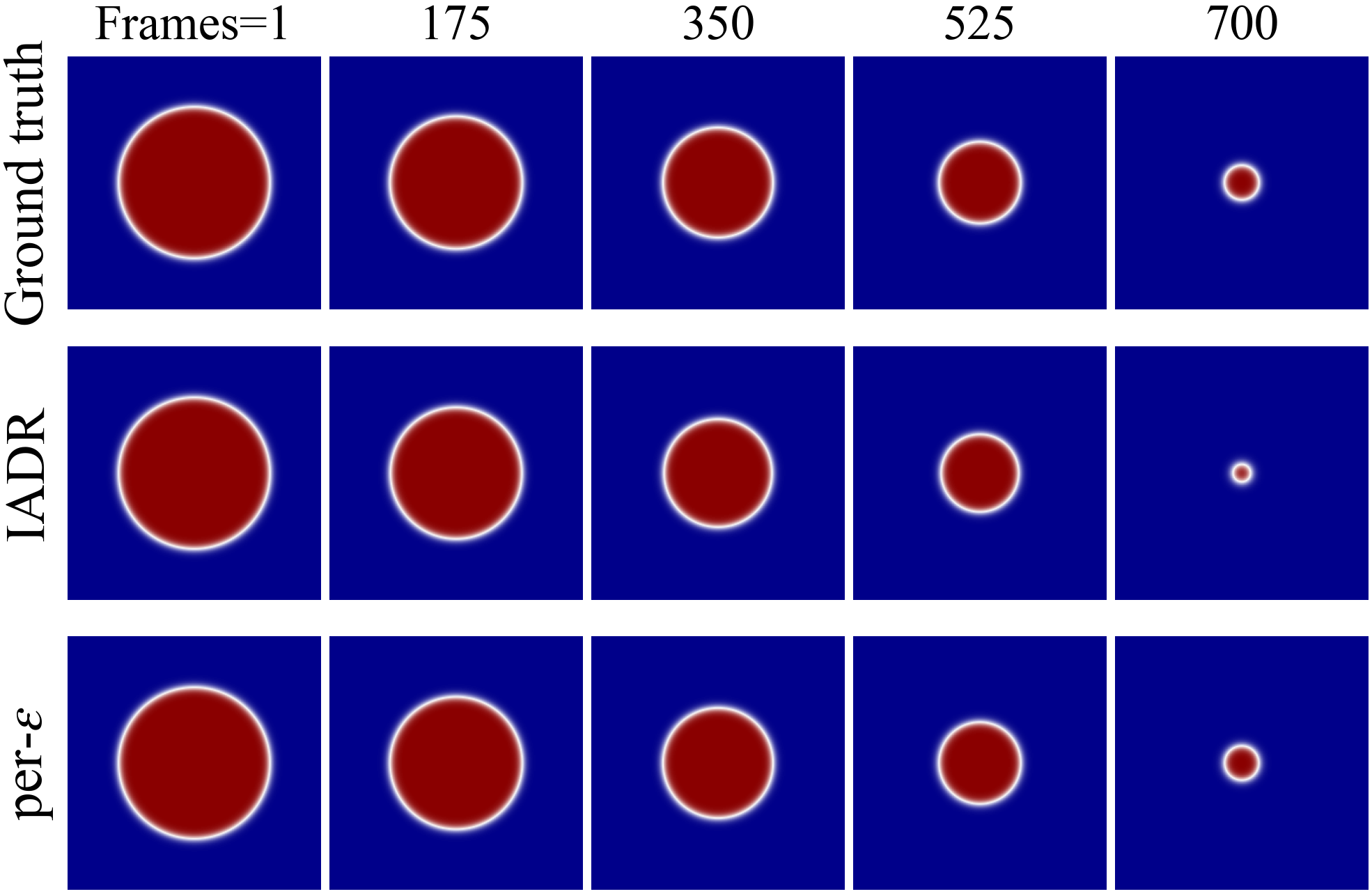}}%
  \end{subfigure}
  \caption{Case-study rollouts; frame indices and model labels are
  marked on each panel.
  \subref{fig:case-circle-mid}~In-distribution circle at
  $\varepsilon = 0.0078$.
  \subref{fig:case-fs}~Joint-OOD five-pointed star at
  $\varepsilon = 0.0069$.
  \subref{fig:case-circle-coarse}~OOD-coarse extrapolation
  $\varepsilon = 0.0093 \notin E$ on a circle.}
  \label{fig:case-study}
\end{figure}

\paragraph{\textsc{IADR} and the per-$\varepsilon$ specialist match
ground truth across the full rollout, while every generic operator
target collapses.}
Figure~\ref{fig:case-circle-mid} shows the in-distribution circle at
$\varepsilon = 0.0078$. \textsc{IADR}
($L^{2}\text{-}T^{\infty} = 22.35 \times 10^{-3}$) and the
per-$\varepsilon$ DR specialist ($6.93 \times 10^{-3}$) coincide
with ground truth at every frame, while U-Net dissolves into
spurious wave patterns by frame $350$ ($959.88 \times 10^{-3}$),
FNO collapses to a low-amplitude blob by frame $175$
($508.08 \times 10^{-3}$), and DISCO fragments into a disconnected
cluster by frame $700$ (diverged,
$L^{2}\text{-}T^{\infty} \approx 4.7 \times 10^{10}$).

\paragraph{\textsc{IADR} retains the per-$\varepsilon$ specialist's
geometric generalisation on a joint-OOD shape.}
Figure~\ref{fig:case-fs} shows the five-pointed star at
$\varepsilon = 0.0069$ (joint OOD). \textsc{IADR}
($L^{2}\text{-}T^{\infty} = 16.73 \times 10^{-3}$),
per-$\varepsilon$ DR ($13.35 \times 10^{-3}$), and ground truth
follow the same evolution from sharp arms through breakup to a small
final residue, while the U-Net target ($720.87 \times 10^{-3}$)
inflates the geometry instead of contracting it. Despite never
seeing a non-circular shape during training, \textsc{IADR} stays
within $1.25\times$ of the per-$\varepsilon$ specialist on this
joint-OOD cell and improves over every generic-operator baseline by
more than an order of magnitude on $L^{2}\text{-}T^{\infty}$
(U-Net $43\times$, FNO $33\times$, DISCO $24\times$ worse), evidence
that the DR-structured target preserves cross-shape generalisation
under hypernetwork conditioning.

\paragraph{\textsc{IADR}'s evolution speed drifts outside the
training range, but the phase-field structure is preserved.}
Figure~\ref{fig:case-circle-coarse} shows the OOD-coarse
extrapolation $\varepsilon = 0.0093 \notin E$. The per-$\varepsilon$
specialist ($L^{2}\text{-}T^{\infty} = 8.27 \times 10^{-3}$) tracks
ground truth throughout, while \textsc{IADR}
($65.95 \times 10^{-3}$) shows a circle at frame $700$ visibly
smaller than the reference, indicating a faster effective evolution
speed. The interface nonetheless
remains a clean phase-field profile, rather than the unstructured
collapse seen for U-Net ($778.98 \times 10^{-3}$), FNO
($510.99 \times 10^{-3}$), or DISCO ($3.1 \times 10^{8}$) on the same
cell, because the DR backbone is hard-coded into the target.

\section{Conclusion}
\label{sec:conclusion}

To meet the practical need to reduce repeated training across interface widths while retaining the phase-field structure and prediction capability of DR operators, we presented \textsc{IADR}, a single $\varepsilon$-conditional neural
operator for the Allen--Cahn equation that pairs a DR splitting target
network with a hypernetwork reading the scalar interface width. A
single \textsc{IADR} matches the per-$\varepsilon$ DR specialists on
in-distribution circles within the prescribed range, retains their cross-shape generalisation on
unseen geometries, and retains their compact DR target structure with millisecond-scale inference. The ablations and comparisons with Direct-$\varepsilon$-DR and specialist weight interpolation further demonstrate the value of full DR conditioning for prediction accuracy and interface-width consistency. The
conditional $K$-step rollout error bound we establish separates one-step
consistency from slice-wise stability.

\begingroup
\section{Limitation and future work}
\label{sec:discussion}
Our discussion of limitations and future work focuses on three aspects: prediction of dynamics involving topological changes with the first-order DR target, architecture optimisation and systematic evaluation across different parameter families, and cross-equation applicability and architecture design.

The first-order DR target may limit prediction performance for dynamics involving topological changes. The dumbbell decomposition experiment in Appendix~\ref{app:more-results} provides preliminary evidence that \textsc{IADR} retains the DR operator's ability to predict dynamics involving a simple topological change. However, both the per-$\varepsilon$ specialists and \textsc{IADR} perform poorly overall in this experiment. Further work is needed to improve the operator design for such dynamics and to conduct a more systematic evaluation.

Our evaluation focuses on the Allen--Cahn equation under fixed numerical settings, with the interface width $\varepsilon$ as the only varying physical parameter. Future work will study joint conditioning on multiple physical parameters and evaluate extensions across different resolutions, time steps, and boundary conditions. We will also explore separating and recombining $\varepsilon$-related and $\varepsilon$-independent features and further study how to incorporate physical constraints into the hypernetwork mechanism, aiming to improve accuracy beyond that of per-$\varepsilon$ specialists.

Although this work focuses only on the Allen--Cahn equation, it also provides a path for cross-equation extensions. The hypernetwork conditioning interface can be retained while adapting the target network and conditioning variables to the corresponding splitting or physical decomposition. When a single MLP generates these weights, its output multiplier $\alpha$ should be selected with the weight scales of the constituent operators in mind. We may also explore using two MLPs to reduce reliance on the output multiplier $\alpha$. For cross-equation extensions, we will first instantiate a target network for a second phase-field PDE, such as the Cahn--Hilliard equation, and independently evaluate the framework's cross-equation transferability.
\endgroup

\begin{ack}
This research is supported by the National Natural Science Foundation of China (No. 12571435 and No. 11971336). The authors declare no competing interests.
\end{ack}

\bibliographystyle{plainnat}
\bibliography{refs}

\appendix

\section{Phase-Field Formulation and DR Splitting Schemes}
\label{app:phase-field-setup}

This appendix expands on the phase-field background summarised in
Section~\ref{sec:method:problem}, derives the first-order DR splitting
scheme of~\eqref{eq:dr-splitting-closed}, and records the higher-order
$K_2 \ast \rho_2$ variant referenced in
Section~\ref{sec:method:approach}.

\subsection{Mean-curvature flow and phase-field representation}
\label{app:math:mcf}

\paragraph{Mean-curvature flow and the Cahn--Hilliard energy.}
A smooth domain $\Omega(t) \subset \mathbb{R}^d$ evolves under mean-curvature
flow if its inward normal velocity satisfies $V_n(t) = H(t)$, where $H(t)$
is the scalar mean curvature of $\partial\Omega(t)$. This is the
$L^2$-gradient flow of the perimeter functional
$P(\Omega(t)) = \int_{\partial\Omega(t)} \mathrm{d}\mathcal{H}^{d-1}$, with
$\mathcal{H}^{d-1}$ the $(d-1)$-dimensional Hausdorff measure. In the
phase-field approach~\citep{chen2002phase,steinbach2009phase}, the
perimeter is approximated (up to a constant) by the Cahn--Hilliard energy
\[
P_\varepsilon(u)
\;=\; \int_{\mathbb{R}^d}
 \left( \varepsilon \frac{|\nabla u|^2}{2}
       + \frac{1}{\varepsilon} W(u) \right) \mathrm{d}x,
\]
with double-well potential $W(s) = \tfrac{1}{2} s^2 (1-s)^2$ fixed
throughout. The energy $\Gamma$-converges to
$P$~\citep{modica1977gamma}, and its $L^2$-gradient flow is
the Allen--Cahn equation~\eqref{eq:allen-cahn}.

\paragraph{Exact phase-field representation and well-prepared data.}
Associated with~\eqref{eq:allen-cahn} is the evolving set
$\Omega_\varepsilon(t) = \{u_\varepsilon(\cdot, t) \geq \tfrac{1}{2}\}$,
with $u_\varepsilon$ the solution under the well-prepared initial data
$u_\varepsilon(x, 0) = q(d(x, \Omega(0)) / \varepsilon)$ for the optimal
profile $q(r) = \tfrac{1}{2}(1 - \tanh(r/2))$. A formal asymptotic expansion
near the interface gives
\[
u_\varepsilon(x, t)
 \;=\; q\!\left(\tfrac{d(x, \Omega_\varepsilon(t))}{\varepsilon}\right)
       + \mathcal{O}(\varepsilon^2),
\]
so that $u_\varepsilon$ stays close to the exact representation
$\mathbf{q}_\varepsilon$ defined in~\eqref{eq:q-representation}. Before the
onset of singularities, the Hausdorff distance between $\Omega_\varepsilon(t)$
and the true MCF evolution $\Omega(t)$ satisfies
\begin{equation}
\label{eq:hausdorff}
\mathrm{dist}_{\mathcal{H}}\big(\Omega(t),\ \Omega_\varepsilon(t)\big)
 \;\leq\; C\, \varepsilon^2 |\log \varepsilon|^2,
\end{equation}
with $C$ depending on the regularity of
$\Omega(t)$~\citep{demottoni1995,chen1992generation}, which is the
tighter-reference statement used in
Section~\ref{sec:method:problem}.

\subsection{First-order DR splitting scheme}
\label{app:math:dr1}

The first-order DR splitting~\citep{bretin2022jcp} is a semi-implicit
discretisation of~\eqref{eq:allen-cahn}: given $\delta t > 0$, one defines
$(u^n)_{n \geq 0}$ by
\begin{equation}
\label{eq:dr-splitting-implicit}
\frac{u^{n+1} - u^n}{\delta t}
 \;=\; \Delta u^{n+1}
      - \frac{1}{\varepsilon^2} W'(u^n),
\end{equation}
which rearranges to
$u^{n+1} = (I - \delta t\, \Delta)^{-1}(u^n - (\delta t / \varepsilon^2) W'(u^n))
 = K_1 \ast \rho_1(u^n)$ with $\rho_1, \widehat{K}_1$ as
in~\eqref{eq:dr-splitting-closed}. The scheme decouples the diffusion and
reaction operators and reaches sub-grid accuracy with only a handful of
parameters per $\varepsilon$, but depends explicitly on $(\delta t,
\varepsilon)$; changing either forces a rebuild.

\subsection{Higher-order DR splitting}
\label{app:higher-order}

The second-order DR splitting scheme~\citep{bretin2022jcp} reads
$u^{n+1} = K_2 \ast \rho_2(u^n)$ with
$\rho_2(s) = s - (\delta t / \varepsilon^2)(W'(s) - \alpha s)$ and
$\widehat{K}_2(\xi) = 1 / (1 + \delta t (4 \pi^2 |\xi|^2 + \alpha / \varepsilon^2))$
for a fixed shift $\alpha > 0$. The corresponding DR splitting network
$\Phi_2$ used by \textsc{IADR} at second order mirrors this scheme by
composing two reaction blocks and two spectral-convolution diffusion blocks
with a skip connection from the input field to the second reaction block,
following the $M = 2$ stage layout~\citep{bretin2022jcp}. The
hypernetwork interface is unchanged: $h_\vartheta(\varepsilon)$ now supplies the
concatenated weight vector of the two $(\rho, K)$ stages together with the
skip coefficients. Fixed-step rollouts evaluate long-horizon prediction. The temporal accuracy of higher-order models will be assessed through time-step refinement experiments in future work.

\section{Proofs and Derivations}
\label{app:proofs}

{This appendix gives (i) a linear non-expansive
special case of Theorem~\ref{thm:rollout-main} in which the exact
propagator is linear, with a tighter telescoping bound; (ii) the formal
version of Theorem~\ref{thm:rollout-main} together with explicit
problem-level and direct hypotheses, and a complete proof; and (iii)
supporting material for the formal version, namely a concrete Lipschitz
estimate on a spectral-gap subspace of the DR splitting backbone, and
the first-order equivalence of the two natural DR splitting orders.}

\subsection{Linear non-expansive special case}
\label{app:proofs:linear}

{We state a linear non-expansive special case of
Theorem~\ref{thm:rollout-main} in which the exact propagator is linear
on a working subspace and the loss is stated through the operator
inverse. In this regime the algebraic telescoping identity
$\mathcal{P}^{K} - \mathcal{Q}^{K} = \sum_{j=0}^{K-1} \mathcal{P}^{j}\, (\mathcal{P} - \mathcal{Q})\, \mathcal{Q}^{K-1-j}$
yields a tighter bound than the recursive estimate of
Theorem~\ref{thm:rollout-main}.

\begin{theorem}[Linear non-expansive special case]
\label{thm:rollout-linear-special}
Let $\widetilde{\mathcal{X}} = \mathbb{R} \times \mathcal{X}$ equipped
with the norm
$\|(r, x)\|_{\widetilde{\mathcal{X}}} := (|r|^{2} + \|x\|_{\mathcal{X}}^{2})^{1/2}$.
Assume the exact propagator
$\mathcal{P}_{\delta t} : \widetilde{\mathcal{X}} \to \mathcal{X}$ is
linear and non-expansive,
\begin{equation}
\label{eq:p-norm}
\|\mathcal{P}_{\delta t}\|_{\widetilde{\mathcal{X}}^{\ast}}
 := \sup_{g \in \widetilde{\mathcal{X}},\, \|g\|_{\widetilde{\mathcal{X}}} = 1}
   \|\mathcal{P}_{\delta t}\, g\|_{\mathcal{X}}
 \;\leq\; 1.
\end{equation}
Let $\mathcal{U} \subseteq \widetilde{\mathcal{X}}$ be a closed linear
subspace with $\mathcal{P}_{\delta t}\, \mathcal{U} \subseteq \mathcal{U}$,
on which $\mathcal{P}_{\delta t}$ is invertible. If the learned linear
approximant $\mathcal{P}^{\mathrm{NN}}_{\theta;\delta t}$ satisfies the
loss bound
\begin{equation}
\label{eq:loss-bound}
\sup_{g \in \mathcal{U},\, \|g\|_{\widetilde{\mathcal{X}}} = 1}
 \big\| \mathcal{P}_{\delta t}^{-1}\, \mathcal{P}^{\mathrm{NN}}_{\theta;\delta t}\, g - g \big\|_{\widetilde{\mathcal{X}}}
 \;\leq\; \epsilon,
\end{equation}
then for every integer $K \geq 1$,
\begin{equation}
\sup_{g \in \mathcal{U},\, \|g\|_{\widetilde{\mathcal{X}}} = 1}
 \big\| \mathcal{P}_{\delta t}^{K}\, g - (\mathcal{P}^{\mathrm{NN}}_{\theta;\delta t})^{K}\, g \big\|_{\mathcal{X}}
 \;\leq\; \epsilon\, K\, (1 + \epsilon)^{K}.
\end{equation}
Moreover, if
$\|\mathcal{P}_{\delta t}\|_{\widetilde{\mathcal{X}}^{\ast}} \leq \eta < 1$
and $\epsilon \leq (1 - \eta)/2$, then
\begin{equation}
\sup_{g \in \mathcal{U},\, \|g\|_{\widetilde{\mathcal{X}}} = 1}
 \big\| \mathcal{P}_{\delta t}^{K}\, g - (\mathcal{P}^{\mathrm{NN}}_{\theta;\delta t})^{K}\, g \big\|_{\mathcal{X}}
 \;\leq\; \epsilon\, K\, \left( \frac{1 + \eta}{2} \right)^{K - 1}.
\end{equation}
\end{theorem}}

\begin{proof}[Proof of Theorem~\ref{thm:rollout-linear-special}]
We work on the unit ball of $\mathcal{U}$,
$\{ g \in \mathcal{U} : \|g\|_{\widetilde{\mathcal{X}}} = 1 \}$,
throughout. For brevity write
$\mathcal{P} := \mathcal{P}_{\delta t}$ and
$\mathcal{P}_{\mathrm{NN}} := \mathcal{P}^{\mathrm{NN}}_{\theta;\delta t}$.

\emph{Step 1: from inverse-form loss to operator error.} For any
$g \in \mathcal{U}$ with $\|g\|_{\widetilde{\mathcal{X}}} = 1$,
\[
\| \mathcal{P}\, g - \mathcal{P}_{\mathrm{NN}}\, g \|_{\mathcal{X}}
 \;=\; \big\| \mathcal{P}\, ( g - \mathcal{P}^{-1}\, \mathcal{P}_{\mathrm{NN}}\, g ) \big\|_{\mathcal{X}}
 \;\leq\; \|\mathcal{P}\|_{\widetilde{\mathcal{X}}^{\ast}} \cdot \epsilon
 \;\leq\; \epsilon,
\]
where the equality uses linearity of $\mathcal{P}$ together with its
invertibility on $\mathcal{U}$, the first inequality uses
non-expansiveness~\eqref{eq:p-norm} and the loss
bound~\eqref{eq:loss-bound}, and the second uses
$\|\mathcal{P}\|_{\widetilde{\mathcal{X}}^{\ast}} \leq 1$. Taking the
supremum,
\begin{equation}
\label{eq:op-err}
\sup_{g \in \mathcal{U},\, \|g\|_{\widetilde{\mathcal{X}}} = 1}
 \| \mathcal{P}\, g - \mathcal{P}_{\mathrm{NN}}\, g \|_{\mathcal{X}}
 \;\leq\; \epsilon.
\end{equation}

\emph{Step 2: bound on the learned operator's norm.} The triangle
inequality, non-expansiveness, and~\eqref{eq:op-err} give
\[
\sup_{g \in \mathcal{U},\, \|g\|_{\widetilde{\mathcal{X}}} = 1}
 \|\mathcal{P}_{\mathrm{NN}}\, g\|_{\mathcal{X}}
 \;\leq\; \|\mathcal{P}\|_{\widetilde{\mathcal{X}}^{\ast}} + \epsilon
 \;\leq\; 1 + \epsilon.
\]

\emph{Step 3: telescoping identity.} For linear operators on the
invariant subspace $\mathcal{U}$,
\[
\mathcal{P}^{K} - \mathcal{P}_{\mathrm{NN}}^{K}
 \;=\; \sum_{j=0}^{K-1} \mathcal{P}^{j}\, (\mathcal{P} - \mathcal{P}_{\mathrm{NN}})\, \mathcal{P}_{\mathrm{NN}}^{K-1-j}.
\]
Applying triangle inequality on the unit ball of $\mathcal{U}$,
sub-multiplicativity of operator norms, and the bounds
$\|\mathcal{P}^{j}\|_{\widetilde{\mathcal{X}}^{\ast}} \leq 1$,
$\|\mathcal{P} - \mathcal{P}_{\mathrm{NN}}\| \leq \epsilon$ (from
\eqref{eq:op-err}),
$\|\mathcal{P}_{\mathrm{NN}}^{K-1-j}\| \leq (1 + \epsilon)^{K-1-j}$ (from
Step 2),
\[
\sup_{g \in \mathcal{U},\, \|g\|_{\widetilde{\mathcal{X}}} = 1}
 \| \mathcal{P}^{K}\, g - \mathcal{P}_{\mathrm{NN}}^{K}\, g \|_{\mathcal{X}}
 \;\leq\; \sum_{j=0}^{K-1} 1 \cdot \epsilon \cdot (1 + \epsilon)^{K-1-j}
 \;\leq\; \epsilon\, K\, (1 + \epsilon)^{K},
\]
which proves the first claim.

\emph{Step 4: strict-contraction branch.} If
$\|\mathcal{P}\|_{\widetilde{\mathcal{X}}^{\ast}} \leq \eta < 1$ and
$\epsilon \leq (1 - \eta)/2$, Step 2 sharpens to
$\|\mathcal{P}_{\mathrm{NN}}\| \leq \eta + \epsilon \leq (1 + \eta)/2$.
Substituting $\|\mathcal{P}^{j}\| \leq \eta^{j}$ and
$\|\mathcal{P}_{\mathrm{NN}}^{K-1-j}\| \leq ((1 + \eta)/2)^{K-1-j}$ into
the telescoping bound, and noting that each term is dominated by
$((1 + \eta)/2)^{K-1}$ since $\eta < (1+\eta)/2$,
\[
\sup_{g \in \mathcal{U},\, \|g\|_{\widetilde{\mathcal{X}}} = 1}
 \| \mathcal{P}^{K}\, g - \mathcal{P}_{\mathrm{NN}}^{K}\, g \|_{\mathcal{X}}
 \;\leq\; \epsilon \sum_{j=0}^{K-1} \eta^{j}\, \left( \frac{1+\eta}{2} \right)^{K-1-j}
 \;\leq\; \epsilon\, K\, \left( \frac{1+\eta}{2} \right)^{K-1},
\]
which proves the second claim. \qedhere
\end{proof}

\subsection{Formal version and proof of Theorem~\ref{thm:rollout-main}}
\label{app:proofs:slice}

{This subsection states the formal slice-wise
version of Theorem~\ref{thm:rollout-main} and gives its complete proof.
The setting is nonlinear (through the reaction $\rho_{1,\varepsilon}$)
and varies with $\varepsilon \in E$, so non-expansiveness and the
inverse-form loss of the linear special case
(Theorem~\ref{thm:rollout-linear-special}) are replaced by slice-wise
Lipschitz stability and a uniform one-step consistency error. This conditional bound describes error propagation under these assumptions.}

\paragraph{Setup.}
We adopt the notation of Section~\ref{sec:method:problem}. Fix
$\varepsilon \in E$ and write
$\mathcal{T}_\varepsilon := \mathcal{T}_{\delta t,\varepsilon}$ and
$\mathcal{N}_\varepsilon := \mathcal{N}_{\theta,\delta t,\varepsilon}$
for the exact and learned one-step propagators at interface width
$\varepsilon$. Both are treated as self-maps of the admissible set
$\mathcal{A}_\varepsilon \subset \mathcal{X}$ under
Assumption~\ref{assum:slice}; the $\varepsilon$-Lipschitz regularity of
$\mathcal{T}_\varepsilon$ (Assumption~\ref{assum:eps-lipschitz}) and the
coverage of the range by $\mu_E$
(Assumption~\ref{assum:coverage}) are not used directly in the
slice-wise proof but justify passing from sampled to uniform control
over $E$.

\paragraph{Problem-level assumptions.}
The three problem-level assumptions summarised in
Section~\ref{sec:method:problem} are formalised as follows.

\begin{assumption}[Coverage of the range]
\label{assum:coverage}
The measure $\mu_E$ is absolutely continuous with respect to Lebesgue
measure on $E$ and has full support, so that a minimiser
of~\eqref{eq:population-loss} controls the loss uniformly over $E$
rather than at a finite set of sampled widths.
\end{assumption}

\begin{assumption}[Slice-wise regularity]
\label{assum:slice}
For every $\varepsilon \in E$, the admissible set $\mathcal{A}_\varepsilon$
is bounded in $\mathcal{X}$ and positively invariant under
$\mathcal{T}_\varepsilon$; that is,
$\mathcal{T}_\varepsilon(\mathcal{A}_\varepsilon) \subseteq
\mathcal{A}_\varepsilon$ up to the horizon $T$.
\end{assumption}

\begin{assumption}[Lipschitz continuity in $\varepsilon$]
\label{assum:eps-lipschitz}
There exists a constant $C_E > 0$ such that for every
$\varepsilon, \varepsilon' \in E$ and every
$\mathbf{q} \in \mathcal{A}_\varepsilon \cap \mathcal{A}_{\varepsilon'}$,
$\| \mathcal{T}_\varepsilon(\mathbf{q})
  - \mathcal{T}_{\varepsilon'}(\mathbf{q}) \|_{L^2(\Omega_x)}
 \leq C_E\, |\varepsilon - \varepsilon'|$.
\end{assumption}

\paragraph{Direct hypotheses of the slice-wise extension.}
The three direct hypotheses are as follows.

\begin{assumption}[Slice-wise Lipschitz stability of the target]
\label{assum:lip-T}
There exists a constant $L_T \geq 0$ such that for every
$\varepsilon \in E$ and every
$\mathbf{q}, \mathbf{q}' \in \mathcal{A}_\varepsilon$,
$\big\| \mathcal{T}_\varepsilon(\mathbf{q})
     - \mathcal{T}_\varepsilon(\mathbf{q}') \big\|_{\mathcal{X}}
 \leq L_T\, \|\mathbf{q} - \mathbf{q}'\|_{\mathcal{X}}$.
\end{assumption}

\begin{assumption}[Slice-wise Lipschitz stability of the learned operator]
\label{assum:lip-N}
There exists a constant $L_N \geq 0$ such that for every
$\varepsilon \in E$ and every
$\mathbf{q}, \mathbf{q}' \in \mathcal{A}_\varepsilon$,
$\big\| \mathcal{N}_\varepsilon(\mathbf{q})
     - \mathcal{N}_\varepsilon(\mathbf{q}') \big\|_{\mathcal{X}}
 \leq L_N\, \|\mathbf{q} - \mathbf{q}'\|_{\mathcal{X}}$.
\end{assumption}

\begin{assumption}[Uniform one-step consistency]
\label{assum:consistency}
There exists $\tau \geq 0$ such that for every $\varepsilon \in E$ and
every $\mathbf{q} \in \mathcal{A}_\varepsilon$,
$\big\| \mathcal{N}_\varepsilon(\mathbf{q})
     - \mathcal{T}_\varepsilon(\mathbf{q}) \big\|_{\mathcal{X}}
 \leq \tau$.
\end{assumption}

\paragraph{On the consistency constant $\tau$.}
The existence of a parameter $\vartheta$ realising $\tau$ in
Assumption~\ref{assum:consistency} is an expressivity statement about
the hypernetwork-parameterised family
$\{\Phi(\cdot; h_\vartheta(\varepsilon))\}_{\varepsilon \in E}$. In
practice it is obtained by a two-step bridging argument: (a) for a
continuous target weight map
$\varepsilon \mapsto w^\star(\varepsilon) \in \mathbb{R}^p$, the
universal approximation theorem for multiple-input
operators~\citep{jin2022mionet} delivers an MLP $h_\vartheta$
approximating $w^\star$ uniformly on
$[\varepsilon_{\min}, \varepsilon_{\max}]$; (b) a Lipschitz-in-weights
estimate
$\|\Phi(\cdot; w) - \Phi(\cdot; \tilde w)\|_{\mathcal{X}}
 \leq C\,\|w - \tilde w\|$ on any bounded input set transfers the
weight-space approximation into operator error. The MIONet theorem does
not cover the hypernetwork parameterisation
$(\varepsilon, \mathbf{q}) \mapsto
 \Phi(\mathbf{q}; h_\vartheta(\varepsilon))$ directly, so we take
Assumption~\ref{assum:consistency} as a hypothesis rather than as a
consequence of MIONet.

\begin{theorem}[Slice-wise $K$-step rollout stability]
\label{thm:rollout-slice}
Under Assumptions~\ref{assum:slice}--\ref{assum:consistency}, set
$L := \max\{L_T, L_N\}$. Then for every integer $K \geq 1$,
\[
\sup_{\varepsilon \in E}
\sup_{\mathbf{q} \in \mathcal{A}_\varepsilon}
\big\| \mathcal{T}_\varepsilon^{K}(\mathbf{q})
      - \mathcal{N}_\varepsilon^{K}(\mathbf{q}) \big\|_{\mathcal{X}}
\;\leq\; \tau \sum_{j=0}^{K-1} L^{j}.
\]
Two regimes follow:
\begin{itemize}
\item[(i)] if $L \geq 1$, the bound specialises to $\tau K\, L^{K-1}$,
which grows linearly in $K$ when $L = 1$ and at geometric rate
$L^{K-1}$ when $L > 1$;
\item[(ii)] if $L < 1$, the geometric series converges and the bound
becomes $\tau / (1 - L)$, uniform in $K$, with residual decaying to
zero at geometric rate $L^{K}$.
\end{itemize}
\end{theorem}

\begin{proof}[Proof of Theorem~\ref{thm:rollout-slice}]
Fix $\varepsilon \in E$, $\mathbf{q} \in \mathcal{A}_\varepsilon$, and
set $u_k := \mathcal{T}_\varepsilon^{k}(\mathbf{q})$,
$v_k := \mathcal{N}_\varepsilon^{k}(\mathbf{q})$,
$e_k := \| u_k - v_k \|_{\mathcal{X}}$.
Assumption~\ref{assum:slice} ensures
$u_k, v_k \in \mathcal{A}_\varepsilon$ for every $k \geq 0$, so
Assumptions~\ref{assum:lip-T}--\ref{assum:consistency} apply pointwise
to the iterates. For every $k \geq 0$,
\begin{align*}
e_{k+1}
 &\;=\; \| \mathcal{T}_\varepsilon(u_k) - \mathcal{N}_\varepsilon(v_k) \|_{\mathcal{X}} \\
 &\;\leq\; \| \mathcal{T}_\varepsilon(u_k) - \mathcal{T}_\varepsilon(v_k) \|_{\mathcal{X}}
      + \| \mathcal{T}_\varepsilon(v_k) - \mathcal{N}_\varepsilon(v_k) \|_{\mathcal{X}}
 \;\leq\; L_T\, e_k + \tau,
\end{align*}
by Assumptions~\ref{assum:lip-T} and~\ref{assum:consistency}. The
symmetric split via Assumptions~\ref{assum:lip-N}
and~\ref{assum:consistency} gives $e_{k+1} \leq L_N\, e_k + \tau$;
either inequality yields $e_{k+1} \leq L\, e_k + \tau$. Iterating from
$e_0 = 0$,
$e_K \leq \tau \sum_{j=0}^{K-1} L^{j}$ for every $K \geq 1$. Taking
suprema over $\mathbf{q} \in \mathcal{A}_\varepsilon$ and
$\varepsilon \in E$ yields the main bound.

For (i), if $L \geq 1$ then $L^{j} \leq L^{K-1}$ for every
$0 \leq j \leq K-1$, so $\tau \sum_{j=0}^{K-1} L^{j} \leq \tau K\, L^{K-1}$.
For (ii), if $L < 1$ the partial sum equals
$\tau (1 - L^{K})/(1 - L) \leq \tau/(1-L)$, with residual
$\tau L^{K}/(1-L) \to 0$ as $K \to \infty$.
\end{proof}

\paragraph{Comparison with Theorem~\ref{thm:rollout-linear-special}.}
{The linear non-expansive special case
(Theorem~\ref{thm:rollout-linear-special}) is recovered from
Theorem~\ref{thm:rollout-main} by setting $E = \{\varepsilon\}$,
$L_T = 1$ (or $\eta$), $L_N = L_T + \epsilon$, $\tau = \epsilon$, with
the algebraic telescoping identity on linear operators tightening the
bound by a constant factor that is unavailable
in the nonlinear setting (Section~\ref{app:proofs:linear} above).}

\subsection{Slice-wise Lipschitz estimate for the DR splitting propagator}
\label{app:proofs:slice-lipschitz}

We record a concrete upper bound on the slice-wise Lipschitz constant
of the DR splitting propagator of
Section~\ref{sec:method:problem}, obtained from a spectral-gap analysis
on a working subspace where the negative Laplacian is coercive.

\begin{lemma}[Slice-wise Lipschitz bound, DR splitting propagator]
\label{lem:dr-slice-lipschitz}
Let $\mathcal{S}_{\delta t,\varepsilon}$ denote the DR splitting operator
of~\eqref{eq:dr-splitting-closed}, and suppose the state field is
constrained to $[0,1]$. Let $Y \subseteq L^2(\Omega_x)$ be a subspace on
which the negative Laplacian has spectral gap $\lambda_1 > 0$ (e.g.\
$Y = L^2(\Omega_x)$ under homogeneous Dirichlet data, or the zero-mean
subspace under periodic or Neumann data). Set
$a := \delta t / \varepsilon^2$ and
$L_\rho(a) := \sup_{s \in [0,1]} |1 - a\, W''(s)|$. Then
\[
\mathrm{Lip}\big(\mathcal{S}_{\delta t,\varepsilon}|_Y\big)
 \;\leq\; \frac{L_\rho(a)}{1 + \delta t\, \lambda_1}.
\]
\end{lemma}

\begin{proof}
Write $\mathcal{S}_{\delta t,\varepsilon}
 = \rho_{1,\varepsilon} \circ (I - \delta t \Delta)^{-1}$ by
\eqref{eq:dr-splitting-closed}, with
$\rho_{1,\varepsilon}(s) = s - (\delta t/\varepsilon^2)\, W'(s)$.

\emph{Step 1 (reaction step).} Since
$\rho_{1,\varepsilon}'(s) = 1 - a\, W''(s)$, the mean value theorem gives
$|\rho_{1,\varepsilon}(u(x)) - \rho_{1,\varepsilon}(v(x))|
 \leq L_\rho(a)\, |u(x) - v(x)|$ almost everywhere on $\Omega_x$ for
$u, v$ taking values in $[0,1]$. Squaring and integrating yields
$\|\rho_{1,\varepsilon}(u) - \rho_{1,\varepsilon}(v)\|_{L^2}
 \leq L_\rho(a)\, \|u - v\|_{L^2}$.

\emph{Step 2 (Helmholtz resolvent).} Expand $f \in Y$ in an orthonormal
eigenbasis $\{\varphi_j\}$ of $-\Delta|_Y$ with eigenvalues
$\mu_j \geq \lambda_1 > 0$, $f = \sum_j c_j \varphi_j$. By the spectral
theorem,
\[
\|(I - \delta t \Delta)^{-1} f\|_{L^2}^2
 \;=\; \sum_j |c_j|^2\, (1 + \delta t\, \mu_j)^{-2}
 \;\leq\; (1 + \delta t\, \lambda_1)^{-2}\, \|f\|_{L^2}^2,
\]
so $\|(I - \delta t \Delta)^{-1}\|_{Y \to Y}
 \leq (1 + \delta t\, \lambda_1)^{-1}$. Moreover, under periodic,
homogeneous Neumann, or homogeneous Dirichlet boundary conditions,
$(I - \delta t \Delta)^{-1}$ is a positive contraction with $\mathbf{1}$
as a fixed point, so $u \in [0,1]$ implies
$(I - \delta t \Delta)^{-1} u \in [0,1]$ as well; the Lipschitz constant
$L_\rho(a)$ of $\rho_{1,\varepsilon}$ taken over $[0,1]$ therefore
remains the relevant bound when $\rho_{1,\varepsilon}$ is applied to
$(I - \delta t \Delta)^{-1} u$.

\emph{Step 3 (composition).} By sub-multiplicativity of Lipschitz
constants on $Y$,
\[
\mathrm{Lip}\big(\mathcal{S}_{\delta t,\varepsilon}|_Y\big)
 \;\leq\; \mathrm{Lip}(\rho_{1,\varepsilon})
          \cdot \|(I - \delta t \Delta)^{-1}\|_{Y \to Y}
 \;\leq\; \frac{L_\rho(a)}{1 + \delta t\, \lambda_1}. \qedhere
\]
\end{proof}

\paragraph{Remark (why the spectral gap is needed).}
On the full $L^2(\Omega_x)$ under periodic boundary data, the zero mode
$\mu_0 = 0$ makes $(I - \delta t \Delta)^{-1}$ only non-expansive.
Combined with
$\rho_{1,\varepsilon}'(\tfrac{1}{2})
 = 1 + \delta t/(2\varepsilon^2) > 1$
for the double-well $W(s) = \tfrac{1}{2} s^2 (1-s)^2$, the Lipschitz
constant on the full space exceeds one.
Restriction to a spectral-gap subspace $Y$ (homogeneous Dirichlet data,
or the zero-mean subspace under periodic or Neumann data) is essential
for a contraction estimate of the DR splitting propagator.

\subsection{First-order equivalence of DR splitting orders}
\label{app:proofs:order-equiv}

The first-order DR splitting admits two natural compositional orders. We
write
\[
\mathcal{S}^{(\rho K)}_{\delta t,\varepsilon}(u)
 \;:=\; K_1 \ast \rho_{1,\varepsilon}(u)
\qquad \text{(reaction-then-diffusion)},
\]
\[
\mathcal{S}^{(K\rho)}_{\delta t,\varepsilon}(u)
 \;:=\; \rho_{1,\varepsilon}(K_1 \ast u)
\qquad \text{(diffusion-then-reaction)}.
\]
The first form arises from the semi-implicit Allen--Cahn step
$(I - \delta t \Delta)\, u^{n+1} = \rho_{1,\varepsilon}(u^{n})$ and is the
form stated by~\citet{bretin2022jcp}; the second form is the order
realised by the reference implementation used in this paper, and matches
the closed-form update~\eqref{eq:dr-splitting-closed} as written. The
following lemma shows that the two orders are first-order equivalent
splittings of the Allen--Cahn semigroup, so the Lipschitz analysis of
Lemma~\ref{lem:dr-slice-lipschitz} and the slice-wise rollout bound of
Theorem~\ref{thm:rollout-slice} apply to either choice.

\begin{lemma}[First-order equivalence of DR splitting orders]
\label{lem:dr-order-equiv}
Let $L_\varepsilon u := \Delta u - \varepsilon^{-2}\, W'(u)$ denote the
Allen--Cahn generator. For every
$u \in C^{2}(\Omega_x; [0,1])$ with bounded Laplacian and every
sufficiently small $\delta t > 0$,
\[
\mathcal{S}^{(\rho K)}_{\delta t,\varepsilon}(u)
 \;=\; u + \delta t\, L_\varepsilon u + O(\delta t^{2}),
\qquad
\mathcal{S}^{(K\rho)}_{\delta t,\varepsilon}(u)
 \;=\; u + \delta t\, L_\varepsilon u + O(\delta t^{2}),
\]
both in the $L^{2}(\Omega_x)$ norm, with $O(\delta t^{2})$ constants
depending on $\|\Delta u\|_{L^{2}}$,
$\|W''\|_{C^{0}([0,1])}$, and $\varepsilon$ but independent of
$\delta t$. In particular,
\[
\big\|\mathcal{S}^{(\rho K)}_{\delta t,\varepsilon}(u)
 - \mathcal{S}^{(K\rho)}_{\delta t,\varepsilon}(u)\big\|_{L^{2}(\Omega_x)}
 \;=\; O(\delta t^{2}).
\]
\end{lemma}

\begin{proof}
Let $T_D := (I - \delta t \Delta)^{-1}$ and $T_R := \rho_{1,\varepsilon}$.
For $u \in C^{2}(\Omega_x)$ with bounded Laplacian, the Helmholtz
resolvent admits the Neumann-series expansion
$T_D u = u + \delta t\, \Delta u + O(\delta t^{2})$ in $L^{2}(\Omega_x)$,
valid for $\delta t < \|\Delta\|^{-1}$ on the working subspace. The
reaction step is exact:
$T_R u = u - (\delta t/\varepsilon^{2})\, W'(u)$.

Composing in the two orders and using a Taylor expansion of $W'$ around
$u$,
\begin{align*}
\mathcal{S}^{(\rho K)}(u)
 &\;=\; T_D(T_R u)
  \;=\; T_R u + \delta t\, \Delta(T_R u) + O(\delta t^{2}) \\
 &\;=\; u - (\delta t/\varepsilon^{2})\, W'(u) + \delta t\, \Delta u + O(\delta t^{2})
  \;=\; u + \delta t\, L_\varepsilon u + O(\delta t^{2}),
\end{align*}
\begin{align*}
\mathcal{S}^{(K\rho)}(u)
 &\;=\; T_R(T_D u)
  \;=\; T_D u - (\delta t/\varepsilon^{2})\, W'(T_D u) \\
 &\;=\; u + \delta t\, \Delta u
   - (\delta t/\varepsilon^{2})\, W'\!\big(u + \delta t\, \Delta u + O(\delta t^{2})\big)
   + O(\delta t^{2}) \\
 &\;=\; u + \delta t\, \Delta u
   - (\delta t/\varepsilon^{2})\, \big(W'(u) + \delta t\, W''(u)\, \Delta u + O(\delta t^{2})\big)
   + O(\delta t^{2}) \\
 &\;=\; u + \delta t\, L_\varepsilon u + O(\delta t^{2}),
\end{align*}
where the cross-term
$-(\delta t^{2}/\varepsilon^{2})\, W''(u)\, \Delta u$ is absorbed into the
$O(\delta t^{2})$ remainder. Subtracting the two expansions yields the
stated equivalence.
\end{proof}

\paragraph{Implication for the rollout analysis.}
Lemma~\ref{lem:dr-slice-lipschitz} bounds the slice-wise Lipschitz
constant of $\mathcal{S}_{\delta t,\varepsilon}$ on a spectral-gap
subspace $Y$ by sub-multiplicativity of Lipschitz constants under
composition; this is symmetric in the order of composition, so the
estimate $L_\rho(a)/(1 + \delta t\, \lambda_1)$ applies to both
$\mathcal{S}^{(\rho K)}$ and $\mathcal{S}^{(K\rho)}$.
Theorem~\ref{thm:rollout-slice} is stated for an abstract slice-wise
propagator $\mathcal{T}_\varepsilon$ with no commitment to its internal
sub-step order, and therefore applies to either splitting order
verbatim.

\section{Implementation Details}
\label{app:impl-details}

\begin{figure}[t]
\centering
\includegraphics[width=\linewidth]{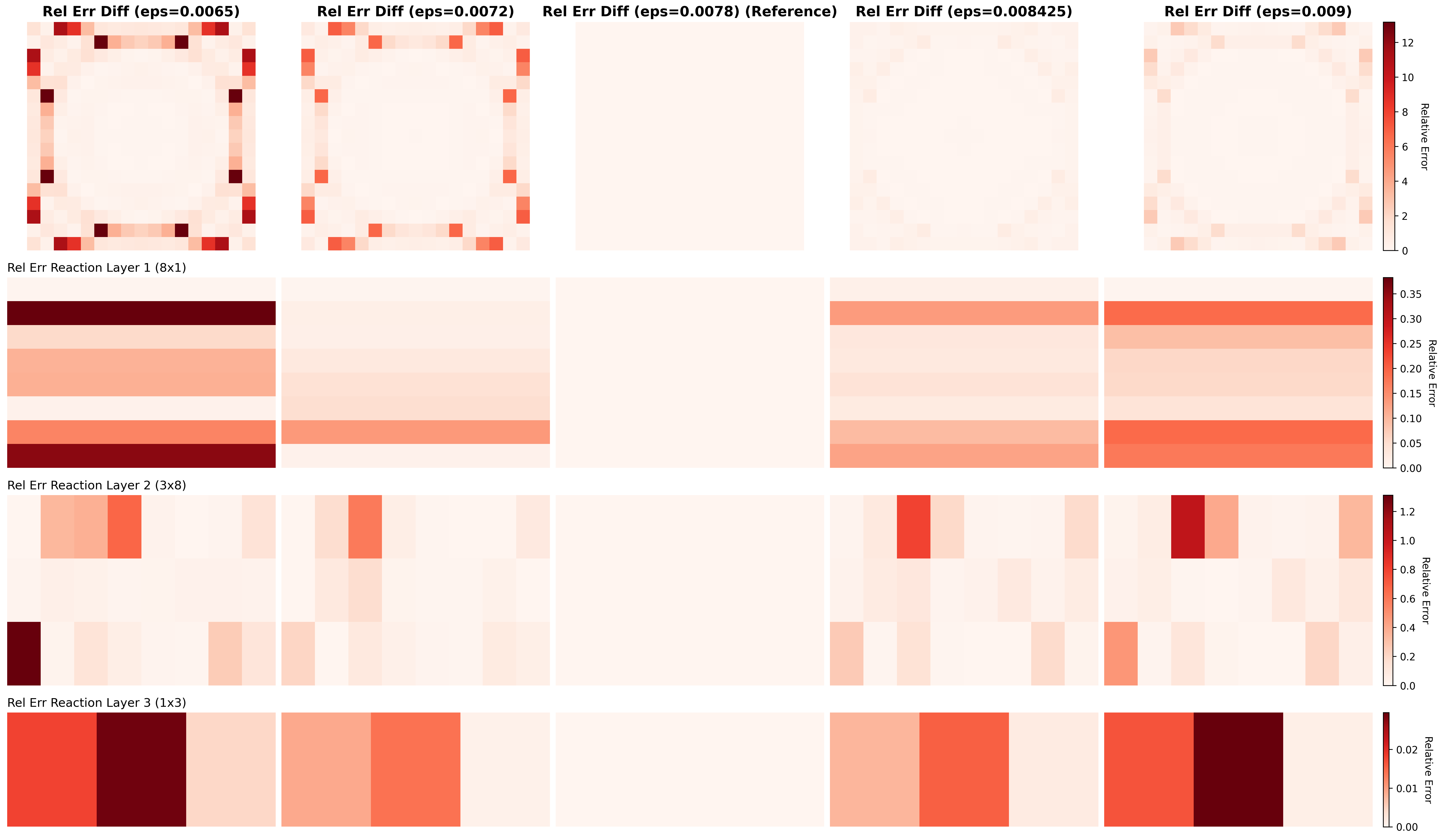}
\caption{Per-$\varepsilon$ DR specialists trained independently at the
five training widths $\{0.0065, 0.0072, 0.0078, 0.0084, 0.0090\}$
exhibit non-trivial weight drift across $\varepsilon$. Row~1 shows the
relative differences of the diffusion-kernel coefficient vectors;
rows~2--4 show the relative differences of the three weight tensors of
the reaction MLP (architecture $1{-}3{-}8{-}1$). All rows take
$\varepsilon = 0.0078$ as the reference; the off-reference columns are
visibly non-zero on structured patterns of coefficients, so a single
shared weight vector cannot match every endpoint. We therefore
parameterise the family by a learned hypernetwork
$h_\vartheta(\varepsilon)$.}
\label{fig:weight-drift}
\end{figure}

\paragraph{IADR architecture.}
The target operator $\Phi$ in~\eqref{eq:dria-phi-decomp} is the first-order
DR splitting network: a spectral convolution parameterised by a
$17 \times 17$ radial kernel initialised at $\widehat{K}_1$
of~\eqref{eq:dr-splitting-closed}, followed by a pointwise reaction MLP of
hidden widths $[8, 3]$ with Gauss activation. The hypernetwork
$h_\vartheta$ is a three-hidden-layer MLP of width $256$ with GELU
activations, reading $\varepsilon$ normalised to $[-1, 1]$ and emitting
directly the flattened weight vector of $\Phi$ ($< 400$ parameters at any
$\varepsilon$). The MLP output is multiplied by a fixed global scalar
factor of $0.1$ before being injected into $\Phi$ via
\texttt{torch.func.functional\_call}, following the weight-injection
mechanism of~\citep{morel2025disco}; $\Phi$ itself carries no trainable
parameters, so the only parameters of \textsc{IADR} are those of
$h_\vartheta$, totalling roughly $160$K.

\paragraph{Baseline architectures.}
The U-Net variant keeps the hypernetwork pipeline of \textsc{IADR} (same
$h_\vartheta$, same \texttt{functional\_call} injection, same global
scalar damping) and replaces $\Phi$ by the four-level U-Net of
\citep{morel2025disco} with base channels $8$ and circular padding
($\sim 200$K target parameters; $\sim 25$M total once the hypernetwork
emits a weight vector of comparable size). The DISCO reproduction follows
the original stack: a space-time Transformer encoder on a $T = 16$ frame
context, hypernetwork head, ODE-integrated single-step prediction, and
DISCO's own U-Net as target; the $T$-frame context consists of $16$ consecutive ground-truth states. At test time, the context starts with $16$ ground-truth states; each autoregressive step removes the earliest state and appends the latest prediction. The Per-$\varepsilon$ DR specialists
share the $\Phi$ topology of \textsc{IADR} but are trained independently at
each width with no hypernetwork. The Blind DR baseline trains the same
$\Phi$ on the pooled five-width data without any $\varepsilon$ input.

\paragraph{Empirical training objective.}
The population loss~\eqref{eq:population-loss} is approximated in training
by an empirical objective in which both integrals over
$(\varepsilon, \mathbf{q})$ are replaced by sums over a finite training set
of $N_{\text{train}}$ pairs $\{(\varepsilon^{(n)}, R_0^{(n)}, t^{(n)})\}_{n=1}^{N_{\text{train}}}$,
and the spatial $L^2(\Omega_x)$ norm by its pixel-grid Riemann sum on the
$H \times W$ grid. With the multipoint unrolling
of~\citet{bretin2022jcp}, each sample contributes $K_{\mathrm{mp}}$
auto-regressive predictions $\Phi^{(k)}(\,\cdot\,; h_\vartheta(\varepsilon))
:= \Phi(\,\cdot\,; h_\vartheta(\varepsilon)) \circ \cdots \circ \Phi(\,\cdot\,; h_\vartheta(\varepsilon))$
($k$-fold composition) compared to the corresponding reference frames, and
the discrete training objective takes the form
\begin{equation}
\label{eq:empirical-loss}
\widehat{\mathcal{L}}(\theta)
\;=\; \frac{1}{N_{\text{train}}\, K_{\mathrm{mp}}}
 \sum_{n=1}^{N_{\text{train}}} \sum_{k=1}^{K_{\mathrm{mp}}}
 \frac{1}{H W}
 \sum_{i,j}
 \Bigl( \Phi^{(k)}\!\bigl(\mathbf{q}^{(n)}; h_\vartheta(\varepsilon^{(n)})\bigr)(i,j)
       - \mathbf{q}_{\varepsilon^{(n)}}\!\bigl(\cdot,\, t^{(n)} + k\delta t\bigr)(i,j)
 \Bigr)^{2},
\end{equation}
where the inner $\frac{1}{H W} \sum_{i,j}(\cdot)^{2}$ is the pixel-mean
squared error implemented in the training script and approximates
$\|\cdot\|_{L^2(\Omega_x)}^{2}$ on the unit square $\Omega_x = [0,1]^{2}$ up
to the Riemann discretisation error of order $\Delta x_{\text{prod}}$. The
configurations reported here use $K_{\mathrm{mp}} = 1$ (single-step
multipoint), with longer unrolls deferred to future runs.

\paragraph{Training schedule and optimiser.}
All trained models share the same base configuration recorded in the
released YAML files: learning rate $1 \times 10^{-4}$, fixed random
seed $0$, and validation every five epochs. The \textsc{IADR},
per-$\varepsilon$ DR, U-Net, FNO, Blind DR, Hyper-D, and Hyper-R
configurations train for $20{,}000$ epochs. DISCO follows its original implementation with a learning rate of $10^{-4}$, $100$ training epochs, and default values for the remaining hyperparameters, without an additional large-scale task-specific hyperparameter search. The optimiser, weight decay, and any learning-rate schedule
are inherited from the \texttt{nnpf} phase-field framework defaults.
The \textsc{IADR} configuration is in \texttt{code/config/IADR.yml}.
To reproduce the reported DISCO training budget, set \texttt{max\_epochs}
to $100$ in \texttt{code/config/FullDisco\_A.yml}. The repository
also provides the configurations for the remaining baselines.

\section{Dataset and Evaluation Details}
\label{app:exp-details}

This appendix collects the dataset construction
(Section~\ref{app:exp-details:datasets}) and the discrete metric
formulas (Section~\ref{app:exp-details:metrics}) of the experimental
suite. Inputs to every model are the raw phase fields
$\mathbf{q}_\varepsilon \in [0, 1]$ on the $256 \times 256$ grid; the
scalar $\varepsilon$ is normalised to $[-1, 1]$ before being read by
the hypernetwork (Appendix~\ref{app:impl-details}). No further
preprocessing or augmentation is applied to the trajectories.

\subsection{Dataset configuration}
\label{app:exp-details:datasets}

Throughout this appendix we denote the standard
training $\varepsilon$ grid by
$E_\text{train} = \{0.0065, 0.0072, 0.0078, 0.0084, 0.0090\}$.

Geometry parameters of the three datasets used in
Section~\ref{sec:exp:setup} are described in the three paragraphs
below; their $\varepsilon$ values, sample counts, and trajectory
frame counts are listed in Table~\ref{tab:dataset-config}.

\paragraph{Circles.} Each circular initial set is a
disk centred near $(0.5, 0.5)$. \emph{(Train / validation.)} Initial
radii are drawn from $R_0 \sim \mathcal{U}[0.05, 0.45]$ with the
centre fixed at $(0.5, 0.5)$. \emph{(Test.)} For every $\varepsilon$
value we draw $100$ independent samples with
$R_0 \sim \mathcal{U}[0.25, 0.35]$ and centre
$(c_x, c_y) \sim \mathcal{U}[0.47, 0.53]^{2}$, so that bootstrap
confidence intervals on circle test cells aggregate over $100$
trajectories rather than a single template.

\paragraph{Circle (DISCO variant).}
DISCO consumes a $16$-frame context per sample (the
\texttt{n\_past}~$=16$ setting of~\citet{morel2025disco}), so for the
DISCO baseline alone we reshape the circle dataset to provide this
context: the train tensor stores $16$ consecutive ground-truth past frames followed by one
target next frame (giving the second dimension $T = 17$), while the
validation tensor stores $16$ past frames followed by six future
targets used for the rollout error metric (giving $T = 22$). The
remaining axes ($N_\varepsilon = 5$ training widths,
$N_\text{sample} = 100$ trajectories per width, single channel,
$256 \times 256$ resolution) and the sampling distribution of $R_0$
are identical to the standard Circle dataset; the test split is
reused unchanged.

\paragraph{Spiral disks.} Each spiral-disk initial set
starts from a disk of base radius $R \in [0.28, 0.32]$ and is deformed
by a spiral-ripple displacement
$a\,\sin(k_r\, r + k_\theta\, \theta)$ in polar coordinates, with
ripple amplitude $a \in [0.06, 0.10]$, integer radial frequency
$k_r \in [9, 11]$, and fixed angular frequency $k_\theta = 5$. The
deformed shape is then rotated by a global angle drawn uniformly from
$[0, 2\pi]$ and translated to the fixed centre $(0.5, 0.5)$.

\paragraph{Five-pointed star.} Each five-pointed-star
initial set is the union of a central disk of radius
$\in [0.10, 0.14]$ and five regular triangular arms (each of outer
radius $\in [0.18, 0.22]$, placed at distance $\in [0.37, 0.43]$ from
the centre, and rotated to the five evenly spaced angular positions
$2\pi i / 5$, $i = 0, \ldots, 4$). The whole figure is then rotated by
a global angle drawn uniformly from $[0, 2\pi]$ and translated to the
fixed centre $(0.5, 0.5)$.

\begin{figure}[h]
\centering
\begin{subfigure}[t]{\linewidth}
  \centering
  \includegraphics[width=0.30\linewidth]{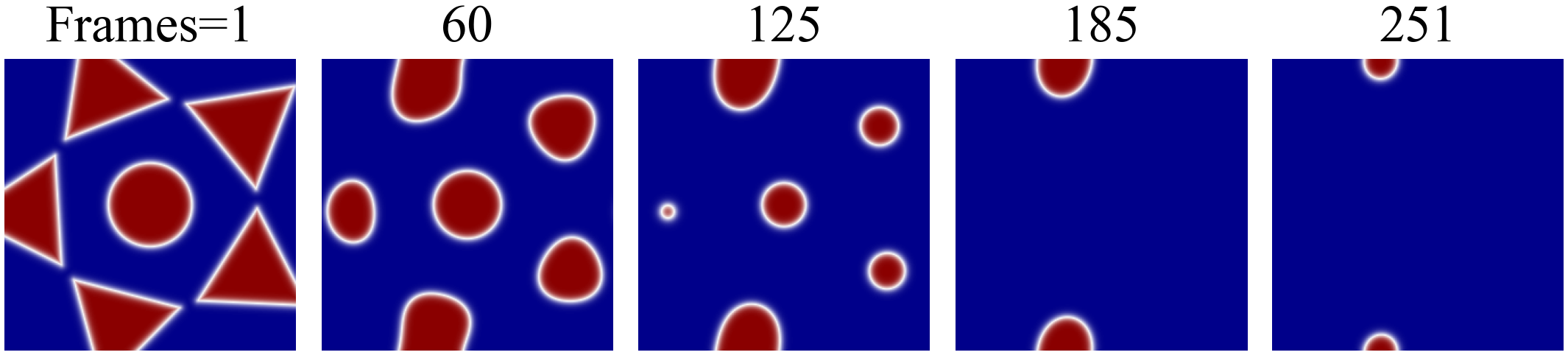}\hfill
  \includegraphics[width=0.30\linewidth]{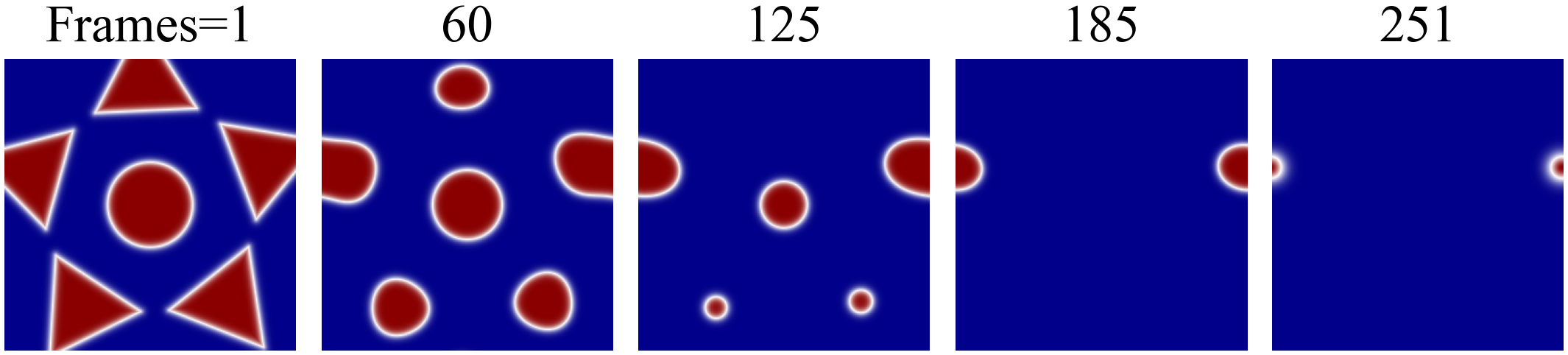}\hfill
  \includegraphics[width=0.30\linewidth]{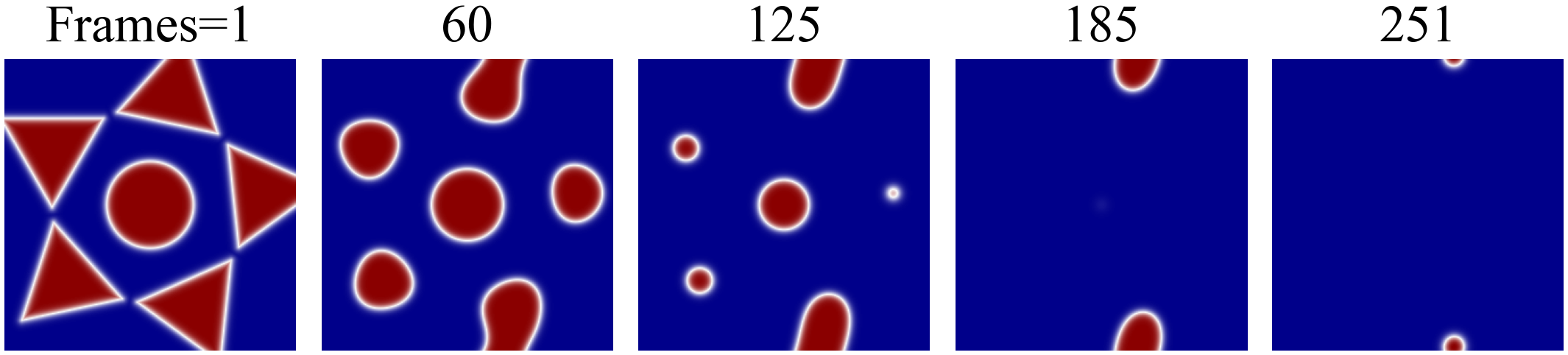}
  \caption{Five-pointed star (\texttt{fs}) initial conditions.}
\end{subfigure}\\[0.5em]
\begin{subfigure}[t]{\linewidth}
  \centering
  \includegraphics[width=0.30\linewidth]{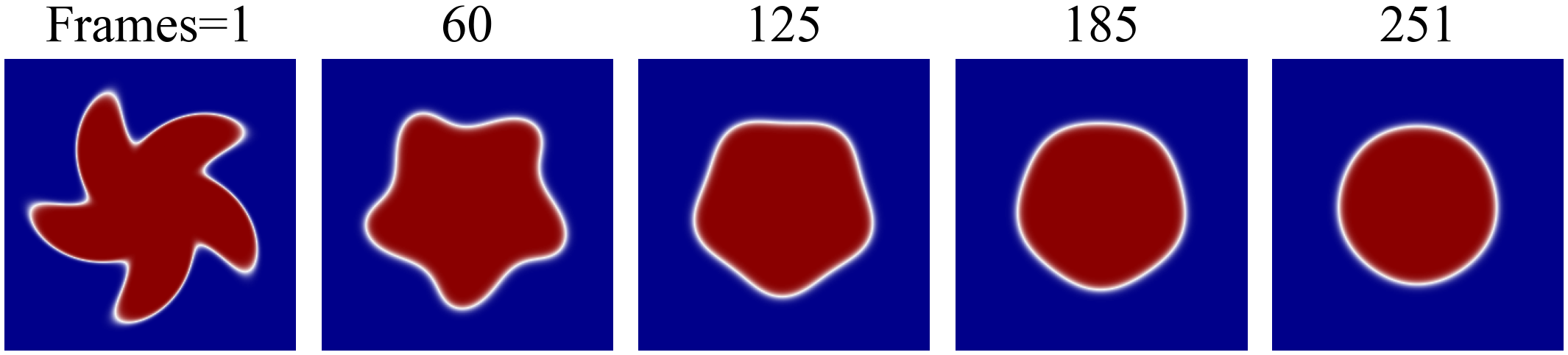}\hfill
  \includegraphics[width=0.30\linewidth]{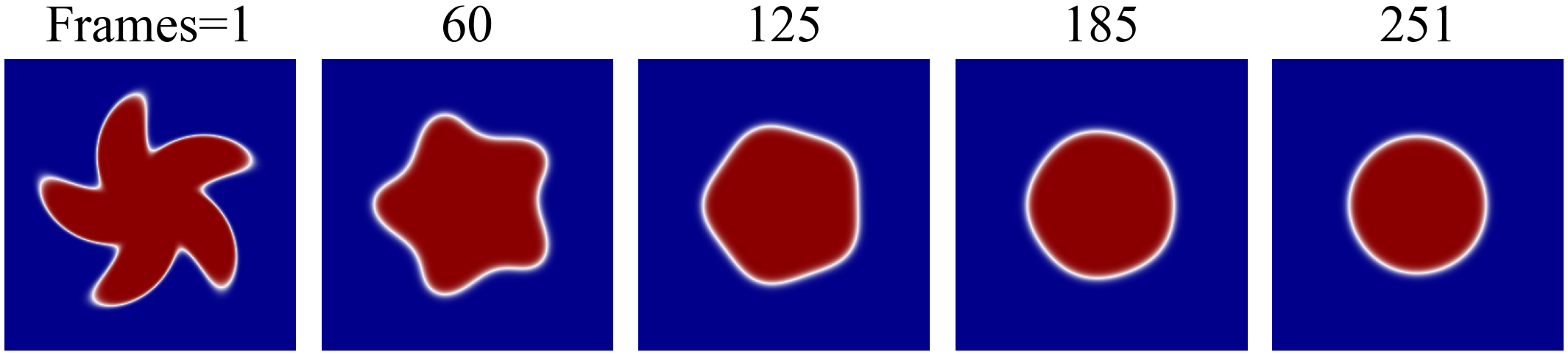}\hfill
  \includegraphics[width=0.30\linewidth]{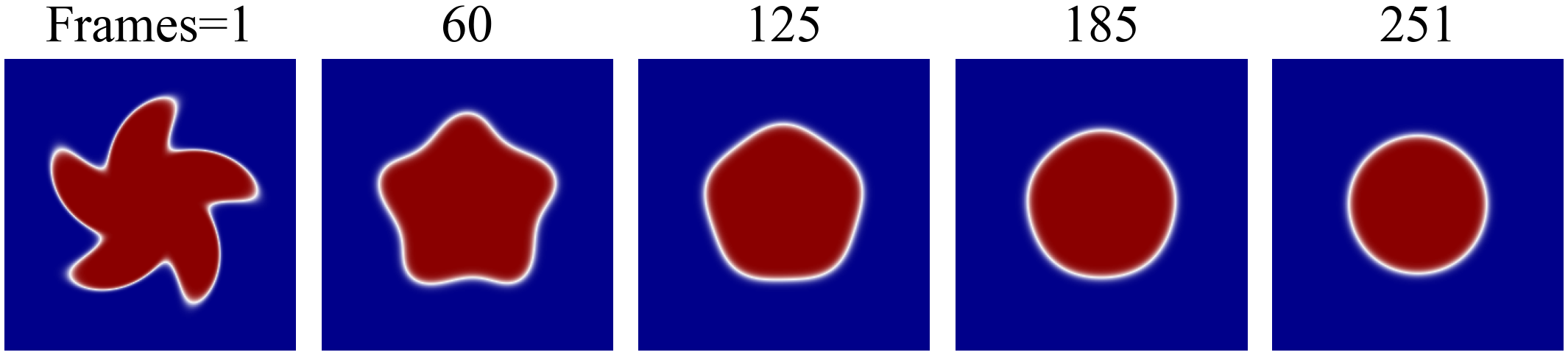}
  \caption{Spiral disk (\texttt{sd}) initial conditions.}
\end{subfigure}
\caption{Sample diversity in the held-out test sets at $\varepsilon = 0.0078$. Three independently sampled initial conditions are shown for each shape family. Samples differ in global rotation, ripple parameters (spiral disk), and arm placement (five-pointed star), so each held-out evaluation aggregates over a non-trivial geometric distribution rather than a single template.}
\label{fig:dataset-panel}
\end{figure}

\begin{table}[h]
\centering
\caption{Dataset configuration. Each \emph{Size} entry
is a 6-tuple $(N_\varepsilon, T, N_\text{sample}, 1, H, W)$, where
$N_\varepsilon$ is the number of sampled $\varepsilon$ values, $T$
is the total number of time frames per sample, $N_\text{sample}$ is
the number of samples per $\varepsilon$, the fourth entry $1$ is the
single channel, and $H \times W$ is the spatial resolution.}
\label{tab:dataset-config}
\scriptsize
\setlength{\tabcolsep}{3pt}
\resizebox{\linewidth}{!}{%
\begin{tabular}{lcccccc}
\toprule
\multirow{2}{*}{Dataset}
 & \multicolumn{2}{c}{Train set}
 & \multicolumn{2}{c}{Validation set}
 & \multicolumn{2}{c}{Test set} \\
\cmidrule(lr){2-3}\cmidrule(lr){4-5}\cmidrule(lr){6-7}
 & $\varepsilon$ & Size
 & $\varepsilon$ & Size
 & $\varepsilon$ & Size \\
\midrule
Circle &
$E_\text{train}$ &
$(5, 2, 100, 1, 256, 256)$ &
$E_\text{train}$ &
$(5, 6, 200, 1, 256, 256)$ &
$E_\text{train} \cup \{0.0069, 0.0081, 0.0093\}$ &
$(8, 700, 100, 1, 256, 256)$ \\
Circle (DISCO variant) &
$E_\text{train}$ &
$(5, 17, 100, 1, 256, 256)$ &
$E_\text{train}$ &
$(5, 22, 100, 1, 256, 256)$ &
\multicolumn{2}{c}{shared with the standard Circle test set} \\
Spiral disks &
--- & --- &
--- & --- &
0.0069, 0.0081, 0.0093 &
$(3, 251, 100, 1, 256, 256)$ \\
Five-pointed star &
--- & --- &
--- & --- &
0.0069, 0.0081, 0.0093 &
$(3, 251, 100, 1, 256, 256)$ \\
\bottomrule
\end{tabular}%
}
\end{table}

\begin{table}[h]
\centering
\caption{Mapping between datasets and tasks.
$\checkmark$ indicates that the dataset is used for the
corresponding split or task; $\times$ indicates it is not.}
\label{tab:dataset-task-map}
\small
\begin{tabular}{lccccc}
\toprule
Dataset &
Train/Val &
Task~1 &
Task~2 &
Task~3 &
Task~4 \\
\midrule
Circle              & $\checkmark$ & $\checkmark$ & $\checkmark$ & $\times$ & $\times$ \\
Spiral disks        & $\times$ & $\times$ & $\times$ & $\checkmark$ & $\checkmark$ \\
Five-pointed star    & $\times$ & $\times$ & $\times$ & $\checkmark$ & $\checkmark$ \\
\bottomrule
\end{tabular}
\end{table}

\subsection{Evaluation metrics}
\label{app:exp-details:metrics}

\paragraph{Notation.}
The formulas below define each metric as the trajectory-mean (over
$N_{\text{total}}$ samples) of a worst-frame quantity per trajectory;
we mark this trajectory mean by an overbar, e.g.,
$\overline{L^{2}_{T^{\infty}}}$,
$\overline{V_{T^{\infty}}}$,
$\overline{\mathrm{AIW}}$. The unbarred shorthands
$L^{2}\text{-}T^{\infty}$, $V\text{-}T^{\infty}$, and AIW used in the
main text and in every results table refer to these same quantities.

Let $\hat u^{(n)}_{t'}, u^{(n)}_{t'} \in \mathbb{R}^{H \times W}$ denote the
predicted and reference phase fields for sample $n$ at time-step index
$t'$, on the $H \times W$ grid with physical cell area
$\Delta x_{\text{prod}} = \Delta x\, \Delta y$. The evaluation window is
$t' \in [t_{\text{start}} + 1,\, t_{\text{end}})$, giving
$T_{\text{eval}} = t_{\text{end}} - t_{\text{start}} - 1$ one-step
predictions per sample across $N_{\text{total}}$ samples.

\paragraph{$L^{2}$-$T^{\infty}$ error.}
The per-frame $L^{2}(\Omega_x)$ error is computed with the physical
cell-area weight,
\[
\bigl\| \hat u^{(n)}_{t'} - u^{(n)}_{t'} \bigr\|_{L^{2}(\Omega_x)}
 \;=\; \sqrt{ \Delta x_{\text{prod}}
              \sum_{i,j} \bigl( \hat u^{(n)}_{t'}(i,j)
                               - u^{(n)}_{t'}(i,j) \bigr)^{2} }.
\]
Taking the worst frame per trajectory and then the sample mean,
\begin{equation}
\label{eq:app:l2-tlinf}
\overline{L^{2}_{T^{\infty}}}
 \;=\; \frac{1}{N_{\text{total}}}
       \sum_{n=1}^{N_{\text{total}}}
       \max_{t'}\,\bigl\| \hat u^{(n)}_{t'} - u^{(n)}_{t'} \bigr\|_{L^{2}(\Omega_x)}.
\end{equation}

\paragraph{Volume-$T^{\infty}$ error.}
The discrete mass of a phase field $u$ is
\[
M[u] \;=\; \Delta x_{\text{prod}} \sum_{i,j} u(i,j)
       \;\approx\; \int_{\Omega_x} u(\mathbf{x})\, \mathrm{d}\mathbf{x},
\]
with per-frame volume error
$\bigl|\, M[\hat u^{(n)}_{t'}] - M[u^{(n)}_{t'}] \,\bigr|$. The reported
scalar is the worst-frame mass deviation, averaged over samples,
\begin{equation}
\label{eq:app:vol-tlinf}
\overline{V_{T^{\infty}}}
 \;=\; \frac{1}{N_{\text{total}}}
       \sum_{n=1}^{N_{\text{total}}}
       \max_{t'}\,\bigl|\, M[\hat u^{(n)}_{t'}] - M[u^{(n)}_{t'}] \,\bigr|.
\end{equation}
Under mean-curvature flow, $M[u_\varepsilon(\cdot,t)]$ decays at the
theoretical rate $A'(t) = -2\pi$ for a single convex curve, so this
metric measures the worst-frame violation of that law.

\paragraph{AIW absolute error.}
The \emph{average interface width} (AIW) of a phase field $u$ is defined
via the discrete coarea formula. Using central differences with
replicate boundary padding,
\[
\partial_x u(i,j)
 = \frac{u(i, j{+}1) - u(i, j{-}1)}{2\, \Delta x},
\quad
\partial_y u(i,j)
 = \frac{u(i{+}1, j) - u(i{-}1, j)}{2\, \Delta y},
\]
we form the regularised gradient magnitude
\[
\|\nabla u\|(i,j)
 \;=\; \sqrt{(\partial_x u)^{2} + (\partial_y u)^{2} + \varepsilon_0},
\qquad \varepsilon_0 = 10^{-30}.
\]
With fixed thresholds $\ell = 0.02$ and $h = 0.98$, define the
interface-band indicator
$\mathbf{1}_{\text{band}}(i,j) = \mathbf{1}\!\left[\ell < u(i,j) < h\right]$
and the band area and band gradient integral
\[
A_{\text{band}}[u] = \Delta x_{\text{prod}}
 \sum_{i,j} \mathbf{1}_{\text{band}}(i,j),
\quad
G_{\text{band}}[u] = \Delta x_{\text{prod}}
 \sum_{i,j} \mathbf{1}_{\text{band}}(i,j)\, \|\nabla u\|(i,j).
\]
The discrete coarea formula yields the interface length estimate
$L_{\text{est}}[u] = G_{\text{band}}[u] / (h - \ell + \varepsilon_0)$,
and the AIW is the band area divided by this length,
\[
\mathrm{AIW}[u]
 \;=\; \frac{A_{\text{band}}[u]}{L_{\text{est}}[u] + \varepsilon_0}.
\]
For a well-prepared $\mathbf{q}_\varepsilon$ profile this equals the
theoretical interface width, proportional to $\varepsilon$ up to a
constant determined by $(\ell, h)$. Fixing a late-time frame index
$t^{\star}$ (default
$t_{\text{start}} + \lfloor 0.8\, (T_{\text{eval}} - 1) \rfloor$), the
per-sample single-frame absolute error is
$\bigl| \mathrm{AIW}[\hat u^{(n)}_{t^{\star}}] - \mathrm{AIW}[u^{(n)}_{t^{\star}}] \bigr|$,
and the reported scalar is its sample mean,
\begin{equation}
\label{eq:app:aiw-single-abs}
\overline{\mathrm{AIW}}
 \;=\; \frac{1}{N_{\text{total}}}
       \sum_{n=1}^{N_{\text{total}}}
       \bigl| \mathrm{AIW}[\hat u^{(n)}_{t^{\star}}]
            - \mathrm{AIW}[u^{(n)}_{t^{\star}}] \bigr|.
\end{equation}

\paragraph{Sample bootstrap.}
Tables~\ref{tab:main}, \ref{tab:main-full}, and~\ref{tab:ablation} report bootstrap means, with $\mathrm{boot\_std}$ additionally shown in Tables~\ref{tab:main-full} and~\ref{tab:ablation}. The $95\%$ percentile CIs are computed but not displayed in these tables. These statistics are computed by non-parametric bootstrap over the $N_{\text{total}}$ per-trajectory metric values $\{ X^{(n)} \}_{n=1}^{N_{\text{total}}}$
(with $X$ one of $\overline{L^{2}_{T^{\infty}}}$,
$\overline{V_{T^{\infty}}}$, $\overline{\mathrm{AIW}}$). For
$b = 1, \ldots, B$ we draw indices
$i_{b,1}, \ldots, i_{b,N_{\text{total}}}$ i.i.d.\ uniformly from
$\{1, \ldots, N_{\text{total}}\}$ with replacement and form the
resampled mean
\[
\bar X^{(b)} \;=\; \frac{1}{N_{\text{total}}}
                   \sum_{k=1}^{N_{\text{total}}} X^{(i_{b,k})}.
\]
The reported point estimate is
$\frac{1}{B}\sum_{b=1}^{B} \bar X^{(b)}$, the standard deviation is
$\mathrm{boot\_std} = \mathrm{std}(\{\bar X^{(b)}\}_{b=1}^{B})$ with
Bessel correction, and the $95\%$ CI is the percentile interval
$[\hat q_{2.5\%},\, \hat q_{97.5\%}]$ of the resampled-mean
distribution. We use $B = 2000$ resamples and a fixed RNG seed of
$42$ for every $(\text{model}, \text{shape}, \varepsilon)$ cell, so
the bootstrap statistics are reproducible from the per-trajectory metric values computed by the evaluation script.

\section{Additional Results}
\label{app:more-results}

\begin{figure}[t]
  \centering
  \includegraphics[width=0.7\linewidth]{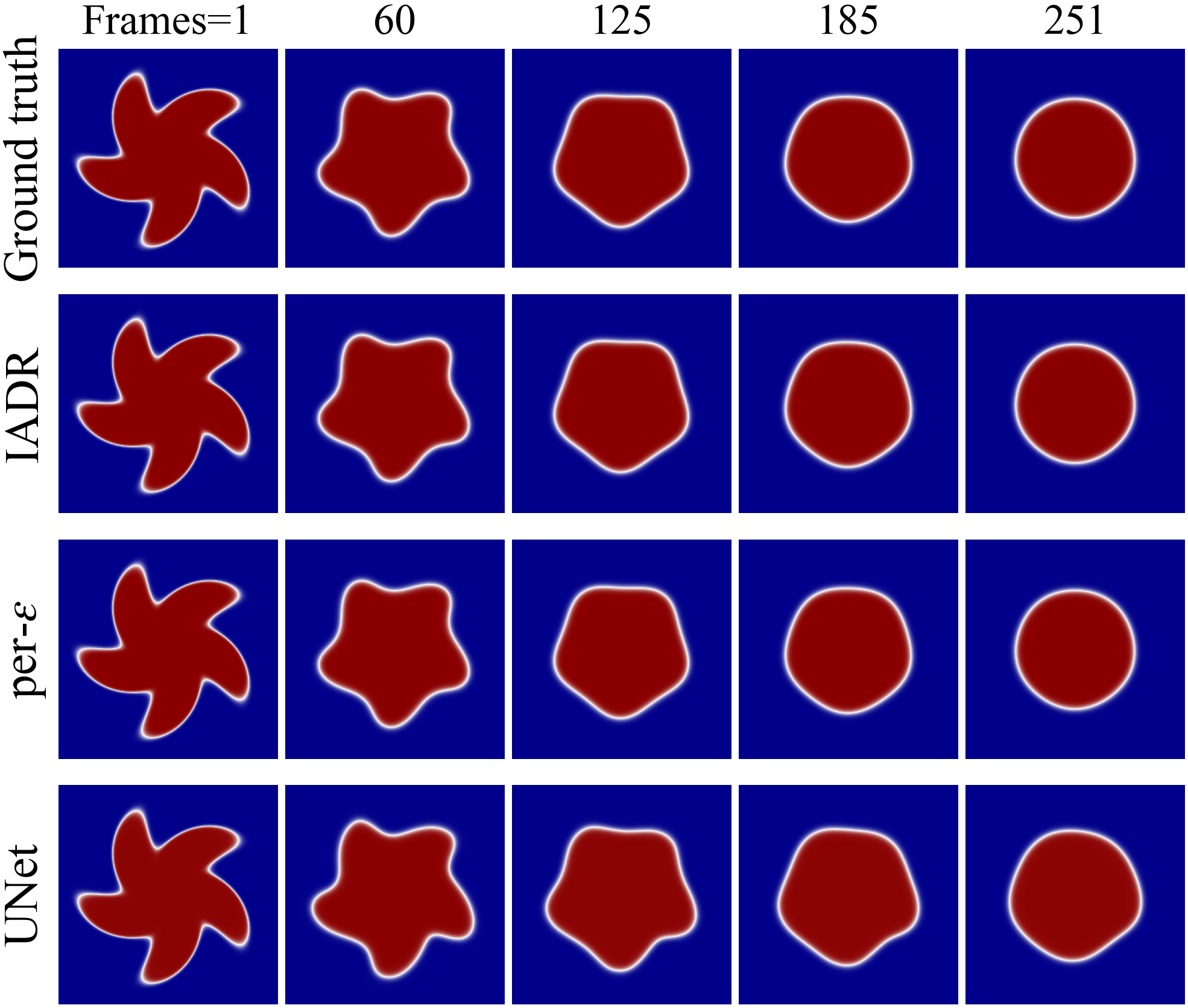}
  \caption{Geometry-OOD spiral disk at $\varepsilon = 0.0078$. Columns
  are frames $\{1, 60, 125, 185, 251\}$; rows are ground truth,
  \textsc{IADR}, the per-$\varepsilon$ DR specialist, and U-Net.}
  \label{fig:case-sd}
\end{figure}

\paragraph{Additional case study: spiral disk geometry-OOD.}
Figure~\ref{fig:case-sd} provides an additional geometry-OOD case:
spiral disk at $\varepsilon = 0.0078$. \textsc{IADR}, per-$\varepsilon$
DR, and ground truth follow the same evolution from spiral arms to a
rounded disk, while the U-Net target rounds the spiral too quickly
and overshoots the contracting disk size at later frames.

\paragraph{Full main-table numbers.}
Table~\ref{tab:main-full} reproduces Table~\ref{tab:main} of the main
text with the full $\mathrm{mean} \pm \mathrm{std}$.
Column scale factors and row grouping are identical; the table is set
in landscape to fit the wider mean $\pm$ std cells.

\begin{sidewaystable}
\centering
\scriptsize
\setlength{\tabcolsep}{2.5pt}
\caption{Main results, full $\mathrm{mean} \pm \mathrm{std}$. Same row
grouping, column scale factors, and per-$\varepsilon$ DR matching as
Table~\ref{tab:main}; smallest mean per (dataset, metric) row is set
in bold; second smallest is underlined. The DISCO entry at
$\varepsilon = 0.0065$ is unavailable.}
\label{tab:main-full}
\resizebox{\linewidth}{!}{%
\begin{tabular}{ll*{15}{c}}
\toprule
 & & \multicolumn{5}{c}{$L^{2}\text{-}T^{\infty}$ ($\times 10^{-3}$)}
 & \multicolumn{5}{c}{$V\text{-}T^{\infty}$ ($\times 10^{-3}$)}
 & \multicolumn{5}{c}{AIW ($\times 10^{-3}$)} \\
\cmidrule(lr){3-7} \cmidrule(lr){8-12} \cmidrule(lr){13-17}
Shape & $\varepsilon$
 & \textsc{IADR} & per-$\varepsilon$ DR & U-Net & FNO & DISCO
 & \textsc{IADR} & per-$\varepsilon$ DR & U-Net & FNO & DISCO
 & \textsc{IADR} & per-$\varepsilon$ DR & U-Net & FNO & DISCO \\
\midrule
\multicolumn{17}{c}{\textit{Task 1: circles at training widths (in-distribution)}} \\
\midrule
 \multirow{5}{*}{Circle} & $0.0065$ & $\underline{33.98} \pm \underline{1.21}$ & $\mathbf{33.75} \pm \mathbf{1.25}$ & $921.67 \pm 0.13$ & $518.77 \pm 5.40$ & diverged & $\underline{3.81} \pm \underline{0.17}$ & $\mathbf{3.79} \pm \mathbf{0.16}$ & $680.39 \pm 0.22$ & $261.42 \pm 6.42$ & diverged & $\mathbf{51.84} \pm \mathbf{49.53}$ & $\underline{51.87} \pm \underline{49.53}$ & $69.46 \pm 48.00$ & $76.61 \pm 46.77$ & diverged \\
  & $0.0072$ & $\underline{27.06} \pm \underline{1.31}$ & $\mathbf{18.89} \pm \mathbf{0.80}$ & $951.33 \pm 0.60$ & $510.70 \pm 6.52$ & $6.7\!\times\!10^{13} \pm 6.6\!\times\!10^{13}$ & $\underline{2.97} \pm \underline{0.10}$ & $\mathbf{2.09} \pm \mathbf{0.12}$ & $709.87 \pm 0.40$ & $269.69 \pm 6.03$ & $7.2\!\times\!10^{12} \pm 7.2\!\times\!10^{12}$ & $\underline{0.34} \pm \underline{0.07}$ & $\mathbf{0.32} \pm \mathbf{0.06}$ & $25.62 \pm 9.18$ & $24.51 \pm 3.49$ & $398.57 \pm 36.02$ \\
  & $0.0078$ & $\underline{22.35} \pm \underline{1.43}$ & $\mathbf{6.93} \pm \mathbf{0.33}$ & $959.88 \pm 0.73$ & $508.08 \pm 5.28$ & $4.7\!\times\!10^{13} \pm 4.6\!\times\!10^{13}$ & $\underline{2.37} \pm \underline{0.03}$ & $\mathbf{0.92} \pm \mathbf{0.08}$ & $728.62 \pm 0.83$ & $269.20 \pm 6.75$ & $5.7\!\times\!10^{12} \pm 5.6\!\times\!10^{12}$ & $\underline{0.48} \pm \underline{0.10}$ & $\mathbf{0.37} \pm \mathbf{0.05}$ & $17.45 \pm 1.56$ & $20.81 \pm 3.53$ & $411.65 \pm 37.82$ \\
  & $0.0084$ & $\underline{14.78} \pm \underline{1.45}$ & $\mathbf{4.53} \pm \mathbf{0.36}$ & $974.31 \pm 0.91$ & $508.02 \pm 5.24$ & $2.5\!\times\!10^{12} \pm 2.4\!\times\!10^{12}$ & $\underline{1.39} \pm \underline{0.08}$ & $\mathbf{0.94} \pm \mathbf{0.09}$ & $745.59 \pm 1.50$ & $264.67 \pm 6.36$ & $2.3\!\times\!10^{11} \pm 2.2\!\times\!10^{11}$ & $\underline{0.54} \pm \underline{0.19}$ & $\mathbf{0.35} \pm \mathbf{0.05}$ & $23.25 \pm 1.21$ & $16.86 \pm 3.49$ & $419.19 \pm 37.11$ \\
  & $0.0090$ & $\underline{11.80} \pm \underline{1.27}$ & $\mathbf{7.36} \pm \mathbf{0.69}$ & $984.91 \pm 1.19$ & $509.65 \pm 5.04$ & $1.6\!\times\!10^{18} \pm 1.6\!\times\!10^{18}$ & $\underline{1.18} \pm \underline{0.07}$ & $\mathbf{0.84} \pm \mathbf{0.04}$ & $755.55 \pm 2.15$ & $268.61 \pm 5.68$ & $3.5\!\times\!10^{17} \pm 3.4\!\times\!10^{17}$ & $\underline{1.00} \pm \underline{0.39}$ & $\mathbf{0.50} \pm \mathbf{0.07}$ & $27.56 \pm 0.98$ & $14.84 \pm 3.41$ & $409.46 \pm 34.45$ \\
\midrule
\multicolumn{17}{c}{\textit{Task 2: circles at non-training widths ($\varepsilon$-OOD)}} \\
\midrule
 \multirow{3}{*}{Circle} & $0.0069$ & $\underline{28.08} \pm \underline{1.22}$ & $\mathbf{25.46} \pm \mathbf{1.04}$ & $937.71 \pm 0.61$ & $508.98 \pm 5.66$ & $1.1\!\times\!10^{10} \pm 5.9\!\times\!10^{9}$ & $\underline{3.08} \pm \underline{0.12}$ & $\mathbf{2.82} \pm \mathbf{0.13}$ & $698.39 \pm 0.29$ & $265.67 \pm 5.89$ & $8.7\!\times\!10^{8} \pm 4.8\!\times\!10^{8}$ & $\mathbf{0.37} \pm \mathbf{0.08}$ & $\underline{0.41} \pm \underline{0.07}$ & $24.17 \pm 9.99$ & $29.90 \pm 2.89$ & $396.72 \pm 34.43$ \\
  & $0.0081$ & $\underline{19.99} \pm \underline{1.49}$ & $\mathbf{9.24} \pm \mathbf{0.40}$ & $968.03 \pm 0.90$ & $500.88 \pm 4.97$ & $7.1\!\times\!10^{9} \pm 3.6\!\times\!10^{9}$ & $\underline{2.07} \pm \underline{0.03}$ & $\mathbf{1.16} \pm \mathbf{0.10}$ & $737.03 \pm 1.13$ & $265.36 \pm 5.98$ & $7.1\!\times\!10^{8} \pm 3.1\!\times\!10^{8}$ & $\underline{0.64} \pm \underline{0.18}$ & $\mathbf{0.50} \pm \mathbf{0.08}$ & $20.79 \pm 1.36$ & $20.59 \pm 3.59$ & $416.88 \pm 38.23$ \\
  & $0.0093$ & $\underline{65.95} \pm \underline{2.25}$ & $\mathbf{8.27} \pm \mathbf{0.75}$ & $778.98 \pm 40.87$ & $510.99 \pm 5.97$ & $3.1\!\times\!10^{11} \pm 1.6\!\times\!10^{11}$ & $\underline{9.97} \pm \underline{0.16}$ & $\mathbf{0.95} \pm \mathbf{0.04}$ & $549.11 \pm 40.59$ & $269.45 \pm 6.06$ & $4.8\!\times\!10^{10} \pm 2.6\!\times\!10^{10}$ & $16.76 \pm 10.10$ & $\mathbf{0.85} \pm \mathbf{0.34}$ & $40.81 \pm 10.95$ & $\underline{12.11} \pm \underline{3.14}$ & $410.05 \pm 35.86$ \\
\midrule
\multicolumn{17}{c}{\textit{Task 3: non-circular geometries at training widths (geometry-OOD)}} \\
\midrule
 \multirow{1}{*}{Five-pointed star} & $0.0078$ & $\underline{21.85} \pm \underline{4.76}$ & $\mathbf{8.80} \pm \mathbf{5.11}$ & $757.35 \pm 1.92$ & $553.48 \pm 4.15$ & $406.12 \pm 2.90$ & $\underline{3.77} \pm \underline{1.09}$ & $\mathbf{1.62} \pm \mathbf{1.19}$ & $603.77 \pm 2.79$ & $229.99 \pm 6.80$ & $224.92 \pm 2.43$ & $\underline{1.98} \pm \underline{1.78}$ & $\mathbf{1.87} \pm \mathbf{1.78}$ & $11.52 \pm 2.66$ & $19.34 \pm 3.24$ & $184.83 \pm 3.96$ \\
 \multirow{1}{*}{Spiral disk} & $0.0078$ & $\underline{4.03} \pm \underline{0.01}$ & $\mathbf{3.65} \pm \mathbf{0.05}$ & $119.04 \pm 3.41$ & $382.58 \pm 6.19$ & $205.19 \pm 1.89$ & $\underline{1.08} \pm \underline{0.01}$ & $\mathbf{0.96} \pm \mathbf{0.02}$ & $33.76 \pm 1.17$ & $86.77 \pm 3.45$ & $16.24 \pm 0.07$ & $0.44 \pm 0.01$ & $\underline{0.35} \pm \underline{0.01}$ & $\mathbf{0.23} \pm \mathbf{0.02}$ & $18.04 \pm 1.16$ & $295.82 \pm 4.42$ \\
\midrule
\multicolumn{17}{c}{\textit{Task 4: non-circular geometries at non-training widths (joint OOD)}} \\
\midrule
 \multirow{2}{*}{Five-pointed star} & $0.0069$ & $\underline{16.73} \pm \underline{5.10}$ & $\mathbf{13.35} \pm \mathbf{4.99}$ & $720.87 \pm 1.69$ & $556.31 \pm 4.54$ & $407.88 \pm 2.87$ & $\underline{2.81} \pm \underline{1.13}$ & $\mathbf{2.25} \pm \mathbf{1.07}$ & $547.74 \pm 2.43$ & $231.35 \pm 7.10$ & $222.74 \pm 1.73$ & $\mathbf{1.70} \pm \mathbf{1.57}$ & $\underline{1.71} \pm \underline{1.58}$ & $7.95 \pm 1.74$ & $20.80 \pm 2.77$ & $180.18 \pm 3.38$ \\
  & $0.0093$ & $\underline{50.26} \pm \underline{5.61}$ & $\mathbf{28.34} \pm \mathbf{6.64}$ & $790.93 \pm 1.14$ & $550.19 \pm 3.96$ & $393.26 \pm 3.05$ & $\underline{10.75} \pm \underline{1.15}$ & $\mathbf{5.23} \pm \mathbf{1.56}$ & $659.78 \pm 1.78$ & $229.31 \pm 6.36$ & $222.89 \pm 2.39$ & $\mathbf{2.33} \pm \mathbf{2.08}$ & $\underline{2.38} \pm \underline{2.07}$ & $9.06 \pm 2.60$ & $13.43 \pm 3.15$ & $186.33 \pm 3.37$ \\
 \multirow{2}{*}{Spiral disk} & $0.0069$ & $\underline{7.71} \pm \underline{0.04}$ & $\mathbf{7.59} \pm \mathbf{0.05}$ & $112.61 \pm 2.53$ & $373.37 \pm 6.56$ & $213.23 \pm 1.59$ & $\underline{1.97} \pm \underline{0.02}$ & $\mathbf{1.95} \pm \mathbf{0.02}$ & $27.20 \pm 0.92$ & $82.75 \pm 3.61$ & $16.06 \pm 0.07$ & $\underline{0.34} \pm \underline{0.02}$ & $0.36 \pm 0.02$ & $\mathbf{0.19} \pm \mathbf{0.02}$ & $24.03 \pm 1.58$ & $286.82 \pm 3.70$ \\
  & $0.0093$ & $\underline{14.16} \pm \underline{0.03}$ & $\mathbf{3.72} \pm \mathbf{0.04}$ & $170.10 \pm 5.60$ & $382.93 \pm 7.89$ & $205.85 \pm 1.56$ & $\underline{4.24} \pm \underline{0.01}$ & $\mathbf{0.64} \pm \mathbf{0.02}$ & $58.13 \pm 2.23$ & $88.09 \pm 3.59$ & $16.00 \pm 0.08$ & $0.56 \pm 0.02$ & $\underline{0.56} \pm \underline{0.01}$ & $\mathbf{0.49} \pm \mathbf{0.03}$ & $8.02 \pm 0.97$ & $293.18 \pm 3.48$ \\
\bottomrule
\end{tabular}%
}
\end{sidewaystable}

\paragraph{Wider range.}
We expand the training interval to $[0.0036,0.0120]$, approximately $3.3$ times the original span, and retain five evenly spaced training widths, including both endpoints. We evaluate circles, five-pointed stars, and spiral disks at $\varepsilon=0.0047$ (unseen interpolation width), $0.0078$ (seen midpoint), and $0.0120$ (seen endpoint). Table~\ref{tab:wide-range} reports the complete comparison with the corresponding per-$\varepsilon$ DR specialists.

At $\varepsilon=0.0120$, \textsc{IADR} has lower errors than the specialists in all nine geometry--metric combinations. Across the nine AIW comparisons in the wider range, \textsc{IADR} has lower error in five. These results show that the original range is not an architectural limit of \textsc{IADR}. The circle AIW errors are higher at the seen endpoint and unseen interpolation width than at the midpoint. The range should be chosen to match the spatial resolution. The experiment supports single-model coverage of a wider predefined, numerically resolvable range included in training, rather than unrestricted extrapolation beyond that range.

\begin{table}[htbp]
\centering
\small
\setlength{\tabcolsep}{2pt}
\caption{Wider-range results. Best result per row in bold.}
\label{tab:wide-range}
\begin{tabular}{ll*{6}{c}}
\toprule
 & & \multicolumn{2}{c}{$L^{2}\text{-}T^{\infty}$ ($\times 10^{-3}$)}
 & \multicolumn{2}{c}{$V\text{-}T^{\infty}$ ($\times 10^{-3}$)}
 & \multicolumn{2}{c}{AIW ($\times 10^{-3}$)} \\
\cmidrule(lr){3-4} \cmidrule(lr){5-6} \cmidrule(lr){7-8}
Shape & $\varepsilon$
 & \textsc{IADR} & per-$\varepsilon$ DR
 & \textsc{IADR} & per-$\varepsilon$ DR
 & \textsc{IADR} & per-$\varepsilon$ DR \\
\midrule
\multicolumn{8}{c}{\textit{Task 1: circles at training widths (in-distribution)}} \\
\midrule
\multirow{2}{*}{Circle} & $0.0078$ & $33.17$ & $\mathbf{7.15}$ & $3.81$ & $\mathbf{0.71}$ & $0.75$ & $\mathbf{0.37}$ \\
 & $0.0120$ & $\mathbf{24.46}$ & $39.72$ & $\mathbf{6.50}$ & $10.87$ & $\mathbf{1.91}$ & $7.35$ \\
\midrule
\multicolumn{8}{c}{\textit{Task 2: circles at non-training widths ($\varepsilon$-OOD)}} \\
\midrule
Circle & $0.0047$ & $96.15$ & $\mathbf{86.47}$ & $23.22$ & $\mathbf{12.19}$ & $7.36$ & $\mathbf{4.55}$ \\
\midrule
\multicolumn{8}{c}{\textit{Task 3: non-circular geometries at training widths (geometry-OOD)}} \\
\midrule
\multirow{2}{*}{Five-pointed star} & $0.0078$ & $46.81$ & $\mathbf{12.10}$ & $8.48$ & $\mathbf{2.44}$ & $1.93$ & $\mathbf{1.90}$ \\
 & $0.0120$ & $\mathbf{40.01}$ & $73.28$ & $\mathbf{7.86}$ & $16.46$ & $\mathbf{0.54}$ & $0.76$ \\
\multirow{2}{*}{Spiral disk} & $0.0078$ & $6.48$ & $\mathbf{2.28}$ & $0.58$ & $\mathbf{0.26}$ & $\mathbf{0.11}$ & $0.30$ \\
 & $0.0120$ & $\mathbf{14.49}$ & $26.61$ & $\mathbf{5.02}$ & $9.13$ & $\mathbf{0.85}$ & $1.34$ \\
\midrule
\multicolumn{8}{c}{\textit{Task 4: non-circular geometries at non-training widths (joint OOD)}} \\
\midrule
Five-pointed star & $0.0047$ & $141.55$ & $\mathbf{86.73}$ & $33.30$ & $\mathbf{15.75}$ & $\mathbf{2.99}$ & $3.26$ \\
Spiral disk & $0.0047$ & $128.90$ & $\mathbf{23.39}$ & $29.96$ & $\mathbf{4.97}$ & $1.39$ & $\mathbf{0.20}$ \\
\bottomrule
\end{tabular}
\end{table}

\paragraph{Dumbbell decomposition.}
The initial configuration consists of two circular lobes joined by a narrow rectangular bridge. During evolution, the bridge breaks and the region separates into multiple connected components. We evaluate the originally trained \textsc{IADR} and per-$\varepsilon$ DR specialists at $\varepsilon=0.0069$ (unseen interpolation width), $0.0078$ (interval midpoint), and $0.0090$ (interval endpoint), without retraining. The results in Table~\ref{tab:dumbbell} indicate that \textsc{IADR} retains geometric evolution capability comparable to that of the per-$\varepsilon$ DR specialists for this simple topological change. The AIW errors further show that \textsc{IADR} continues to control the phase-field interface width.

\begin{table}[htbp]
\centering
\small
\caption{Dumbbell decomposition using the originally trained models.}
\label{tab:dumbbell}
\setlength{\tabcolsep}{5pt}
\begin{tabular}{c*{6}{c}}
\toprule
 & \multicolumn{2}{c}{$L^{2}\text{-}T^{\infty}$ ($\times10^{-2}$)}
 & \multicolumn{2}{c}{$V\text{-}T^{\infty}$ ($\times10^{-2}$)}
 & \multicolumn{2}{c}{AIW ($\times10^{-3}$)} \\
\cmidrule(lr){2-3}\cmidrule(lr){4-5}\cmidrule(lr){6-7}
$\varepsilon$ & \textsc{IADR} & Specialist & \textsc{IADR} & Specialist & \textsc{IADR} & Specialist \\
\midrule
$0.0069$ & $6.548$ & $\mathbf{3.801}$ & $1.048$ & $\mathbf{0.5869}$ & $0.8420$ & $\mathbf{0.4394}$ \\
$0.0078$ & $10.47$ & $\mathbf{7.542}$ & $2.167$ & $\mathbf{1.696}$ & $2.363$ & $\mathbf{2.198}$ \\
$0.0090$ & $10.52$ & $\mathbf{8.329}$ & $2.060$ & $\mathbf{1.634}$ & $2.233$ & $\mathbf{2.052}$ \\
\bottomrule
\end{tabular}
\end{table}

\section{Ablation Results}
\label{app:ablation-results}

This appendix reports the per-cell numbers for the Blind DR, Hyper-D
and Hyper-R ablations: variants of \textsc{IADR} that remove or
restrict $\varepsilon$-conditioning while keeping the DR splitting
target fixed. \textbf{Blind DR} replaces the hypernetwork with a
single $\varepsilon$-agnostic weight vector trained on the pooled
five-width data. \textbf{Hyper-D} and \textbf{Hyper-R} restrict the
hypernetwork to emit only the diffusion-block weights $w_D \in
\mathbb{R}^{K_D^{2}}$ or only the reaction-block weights $w_R \in
\mathbb{R}^{p_R}$ respectively, with the non-emitted block taking a
single $\varepsilon$-agnostic vector jointly trained with
$h_\vartheta$. On circles, we also compare Direct-$\varepsilon$-DR, which uses parameter concatenation, at $\varepsilon=0.0069$, $0.0078$ and $0.0090$, and direct interpolation of independently trained specialist weights (without further training) at the unseen widths $0.0069$ and $0.0081$. Tables~\ref{tab:direct-eps-comparison} and~\ref{tab:weight-interpolation} report these results; \textsc{IADR} has lower errors on all three metrics in these comparisons. Row grouping and column scale factors mirror
Table~\ref{tab:main}; smallest mean per (dataset, metric) row across
the four ablation columns is set in bold; second smallest is
underlined.

\begin{sidewaystable}
\centering
\scriptsize
\setlength{\tabcolsep}{2.5pt}
\caption{Blind DR / Hyper-D / Hyper-R ablations, full $\mathrm{mean} \pm
\mathrm{std}$ at the same (shape, $\varepsilon$) cells as
Table~\ref{tab:main}.}
\label{tab:ablation}
\resizebox{\linewidth}{!}{%
\begin{tabular}{ll*{12}{c}}
\toprule
 & & \multicolumn{4}{c}{$L^{2}\text{-}T^{\infty}$ ($\times 10^{-3}$)}
 & \multicolumn{4}{c}{$V\text{-}T^{\infty}$ ($\times 10^{-3}$)}
 & \multicolumn{4}{c}{AIW ($\times 10^{-3}$)} \\
\cmidrule(lr){3-6} \cmidrule(lr){7-10} \cmidrule(lr){11-14}
Shape & $\varepsilon$
 & \textsc{IADR} & Blind DR & Hyper-D & Hyper-R
 & \textsc{IADR} & Blind DR & Hyper-D & Hyper-R
 & \textsc{IADR} & Blind DR & Hyper-D & Hyper-R \\
\midrule
\multicolumn{14}{c}{\textit{Task 1: circles at training widths (in-distribution)}} \\
\midrule
 \multirow{5}{*}{Circle} & $0.0065$ & $\underline{33.98} \pm \underline{1.21}$ & $322.62 \pm 6.85$ & $177.20 \pm 3.78$ & $\mathbf{10.22} \pm \mathbf{0.86}$ & $\underline{3.81} \pm \underline{0.17}$ & $114.81 \pm 2.49$ & $40.18 \pm 0.70$ & $\mathbf{1.03} \pm \mathbf{0.06}$ & $\underline{51.84} \pm \underline{49.53}$ & $82.20 \pm 48.14$ & $53.23 \pm 48.15$ & $\mathbf{51.81} \pm \mathbf{49.53}$ \\
  & $0.0072$ & $\underline{27.06} \pm \underline{1.31}$ & $324.67 \pm 6.64$ & $87.07 \pm 5.72$ & $\mathbf{18.66} \pm \mathbf{0.74}$ & $\underline{2.97} \pm \underline{0.10}$ & $117.41 \pm 2.50$ & $13.77 \pm 0.70$ & $\mathbf{2.25} \pm \mathbf{0.12}$ & $\mathbf{0.34} \pm \mathbf{0.07}$ & $36.84 \pm 0.96$ & $\underline{5.33} \pm \underline{2.79}$ & $72.38 \pm 69.27$ \\
  & $0.0078$ & $\mathbf{22.35} \pm \mathbf{1.43}$ & $327.27 \pm 6.66$ & $65.00 \pm 1.12$ & $\underline{33.13} \pm \underline{1.50}$ & $\mathbf{2.37} \pm \mathbf{0.03}$ & $120.55 \pm 2.61$ & $9.47 \pm 0.58$ & $\underline{4.13} \pm \underline{0.11}$ & $\mathbf{0.48} \pm \mathbf{0.10}$ & $41.32 \pm 1.20$ & $18.65 \pm 18.13$ & $\underline{7.85} \pm \underline{7.12}$ \\
  & $0.0084$ & $\mathbf{14.78} \pm \mathbf{1.45}$ & $341.33 \pm 6.78$ & $158.59 \pm 1.67$ & $\underline{48.62} \pm \underline{1.94}$ & $\mathbf{1.39} \pm \mathbf{0.08}$ & $133.54 \pm 2.80$ & $33.21 \pm 0.80$ & $\underline{6.76} \pm \underline{0.15}$ & $\mathbf{0.54} \pm \mathbf{0.19}$ & $46.01 \pm 1.44$ & $22.55 \pm 8.73$ & $\underline{3.36} \pm \underline{2.66}$ \\
  & $0.0090$ & $\mathbf{11.80} \pm \mathbf{1.27}$ & $360.34 \pm 6.02$ & $205.08 \pm 2.21$ & $\underline{54.13} \pm \underline{2.13}$ & $\mathbf{1.18} \pm \mathbf{0.07}$ & $150.91 \pm 2.48$ & $51.18 \pm 1.17$ & $\underline{7.95} \pm \underline{0.15}$ & $\mathbf{1.00} \pm \mathbf{0.39}$ & $50.29 \pm 1.70$ & $46.32 \pm 15.83$ & $\underline{3.53} \pm \underline{2.63}$ \\
\midrule
\multicolumn{14}{c}{\textit{Task 2: circles at non-training widths ($\varepsilon$-OOD)}} \\
\midrule
 \multirow{3}{*}{Circle} & $0.0069$ & $\mathbf{28.08} \pm \mathbf{1.22}$ & $323.69 \pm 6.76$ & $131.63 \pm 4.96$ & $\underline{36.79} \pm \underline{1.16}$ & $\mathbf{3.08} \pm \mathbf{0.12}$ & $116.35 \pm 2.57$ & $25.37 \pm 0.72$ & $\underline{4.58} \pm \underline{0.22}$ & $\mathbf{0.37} \pm \mathbf{0.08}$ & $34.65 \pm 0.84$ & $5.02 \pm 2.59$ & $\underline{2.40} \pm \underline{2.00}$ \\
  & $0.0081$ & $\mathbf{19.99} \pm \mathbf{1.49}$ & $331.30 \pm 6.60$ & $115.58 \pm 1.44$ & $\underline{48.97} \pm \underline{1.83}$ & $\mathbf{2.07} \pm \mathbf{0.03}$ & $124.22 \pm 2.69$ & $20.34 \pm 0.62$ & $\underline{6.75} \pm \underline{0.17}$ & $\mathbf{0.64} \pm \mathbf{0.18}$ & $43.61 \pm 1.32$ & $11.65 \pm 4.15$ & $\underline{3.03} \pm \underline{2.32}$ \\
  & $0.0093$ & $\underline{65.95} \pm \underline{2.25}$ & $366.49 \pm 5.91$ & $212.02 \pm 2.26$ & $\mathbf{39.37} \pm \mathbf{2.08}$ & $\underline{9.97} \pm \underline{0.16}$ & $156.89 \pm 2.62$ & $54.44 \pm 1.22$ & $\mathbf{5.37} \pm \mathbf{0.06}$ & $\underline{16.76} \pm \underline{10.10}$ & $52.66 \pm 1.84$ & $51.79 \pm 18.82$ & $\mathbf{12.11} \pm \mathbf{10.64}$ \\
\midrule
\multicolumn{14}{c}{\textit{Task 3: non-circular geometries at training widths (geometry-OOD)}} \\
\midrule
 \multirow{1}{*}{Five-pointed star} & $0.0078$ & $21.85 \pm 4.76$ & $327.30 \pm 3.90$ & $\underline{16.95} \pm \underline{4.77}$ & $\mathbf{12.33} \pm \mathbf{5.12}$ & $3.77 \pm 1.09$ & $145.59 \pm 2.67$ & $\underline{2.95} \pm \underline{1.11}$ & $\mathbf{2.75} \pm \mathbf{1.21}$ & $\underline{1.98} \pm \underline{1.78}$ & $38.82 \pm 1.09$ & $\mathbf{1.92} \pm \mathbf{1.79}$ & $2.04 \pm 1.77$ \\
 \multirow{1}{*}{Spiral disk} & $0.0078$ & $\mathbf{4.03} \pm \mathbf{0.01}$ & $331.22 \pm 2.63$ & $26.29 \pm 0.22$ & $\underline{7.00} \pm \underline{0.02}$ & $\mathbf{1.08} \pm \mathbf{0.01}$ & $14.53 \pm 0.65$ & $7.19 \pm 0.09$ & $\underline{1.99} \pm \underline{0.01}$ & $\underline{0.44} \pm \underline{0.01}$ & $41.47 \pm 0.03$ & $\mathbf{0.29} \pm \mathbf{0.01}$ & $0.44 \pm 0.02$ \\
\midrule
\multicolumn{14}{c}{\textit{Task 4: non-circular geometries at non-training widths (joint OOD)}} \\
\midrule
 \multirow{2}{*}{Five-pointed star} & $0.0069$ & $\underline{16.73} \pm \underline{5.10}$ & $325.14 \pm 3.81$ & $163.30 \pm 1.74$ & $\mathbf{15.52} \pm \mathbf{5.33}$ & $\underline{2.81} \pm \underline{1.13}$ & $141.47 \pm 2.54$ & $43.35 \pm 0.60$ & $\mathbf{2.07} \pm \mathbf{1.24}$ & $\mathbf{1.70} \pm \mathbf{1.57}$ & $33.16 \pm 0.54$ & $4.76 \pm 2.24$ & $\underline{1.74} \pm \underline{1.57}$ \\
  & $0.0093$ & $\underline{50.26} \pm \underline{5.61}$ & $348.38 \pm 3.65$ & $177.55 \pm 2.80$ & $\mathbf{24.82} \pm \mathbf{5.34}$ & $\underline{10.75} \pm \underline{1.15}$ & $166.93 \pm 2.71$ & $54.46 \pm 1.41$ & $\mathbf{5.70} \pm \mathbf{1.42}$ & $\mathbf{2.33} \pm \mathbf{2.08}$ & $51.70 \pm 1.30$ & $270.10 \pm 136.07$ & $\underline{2.75} \pm \underline{2.05}$ \\
 \multirow{2}{*}{Spiral disk} & $0.0069$ & $\mathbf{7.71} \pm \mathbf{0.04}$ & $332.76 \pm 2.19$ & $46.14 \pm 0.47$ & $\underline{10.74} \pm \underline{0.07}$ & $\mathbf{1.97} \pm \mathbf{0.02}$ & $19.20 \pm 0.70$ & $3.23 \pm 0.15$ & $\underline{2.86} \pm \underline{0.03}$ & $\underline{0.34} \pm \underline{0.02}$ & $34.49 \pm 0.02$ & $\mathbf{0.23} \pm \mathbf{0.02}$ & $0.39 \pm 0.02$ \\
  & $0.0093$ & $\underline{14.16} \pm \underline{0.03}$ & $334.50 \pm 2.03$ & $73.80 \pm 0.46$ & $\mathbf{6.29} \pm \mathbf{0.02}$ & $\underline{4.24} \pm \underline{0.01}$ & $14.02 \pm 0.81$ & $21.17 \pm 0.06$ & $\mathbf{1.94} \pm \mathbf{0.00}$ & $\mathbf{0.56} \pm \mathbf{0.02}$ & $53.15 \pm 0.04$ & $2.00 \pm 0.02$ & $\underline{0.94} \pm \underline{0.01}$ \\
\bottomrule
\end{tabular}%
}
\end{sidewaystable}

Although Hyper-R appears competitive on most cells of
Table~\ref{tab:ablation}, its AIW at $\varepsilon = 0.0072$ rises to
$72.38 \times 10^{-3}$, two orders of magnitude above \textsc{IADR}'s
$0.34 \times 10^{-3}$ and even higher than Blind DR ($36.84$) and
Hyper-D ($5.33$) at the same cell. The width $\varepsilon = 0.0072$ is
neither the boundary of the training interval ($\varepsilon = 0.0065$)
nor an out-of-range value ($\varepsilon = 0.0093$); it is a width the
model was trained on, in the interior of the training set. This result shows the value of joint diffusion and reaction conditioning for interface-width consistency and, together with the Direct-$\varepsilon$-DR and specialist weight interpolation comparisons in Tables~\ref{tab:direct-eps-comparison} and~\ref{tab:weight-interpolation}, supports full DR conditioning.

\begin{table}[htbp]
\centering
\small
\caption{Comparison with Direct-$\varepsilon$-DR on circles. All errors are in units of $\times10^{-3}$.}
\label{tab:direct-eps-comparison}
\begin{tabular}{clccc}
\toprule
$\varepsilon$ & Model & $L^{2}\text{-}T^{\infty}$ & $V\text{-}T^{\infty}$ & AIW \\
\midrule
\multirow{2}{*}{$0.0069$} & \textsc{IADR} & $\mathbf{28.08}$ & $\mathbf{3.08}$ & $\mathbf{0.37}$ \\
 & Direct-$\varepsilon$-DR & $34.07$ & $4.29$ & $2.06$ \\
\midrule
\multirow{2}{*}{$0.0078$} & \textsc{IADR} & $\mathbf{22.35}$ & $\mathbf{2.37}$ & $\mathbf{0.48}$ \\
 & Direct-$\varepsilon$-DR & $33.50$ & $5.37$ & $8.03$ \\
\midrule
\multirow{2}{*}{$0.0090$} & \textsc{IADR} & $\mathbf{11.80}$ & $\mathbf{1.18}$ & $\mathbf{1.00}$ \\
 & Direct-$\varepsilon$-DR & $47.24$ & $7.74$ & $13.92$ \\
\bottomrule
\end{tabular}
\end{table}

\begin{table}[htbp]
\centering
\small
\caption{Comparison with specialist weight interpolation on circles at the unseen widths $\varepsilon=0.0069$ and $0.0081$. All errors are in units of $\times10^{-3}$.}
\label{tab:weight-interpolation}
\begin{tabular}{clccc}
\toprule
$\varepsilon$ & Model & $L^{2}\text{-}T^{\infty}$ & $V\text{-}T^{\infty}$ & AIW \\
\midrule
\multirow{2}{*}{$0.0069$} & \textsc{IADR} & $\mathbf{28.08}$ & $\mathbf{3.08}$ & $\mathbf{0.37}$ \\
 & Specialist weight interpolation & $68.71$ & $9.56$ & $23.71$ \\
\midrule
\multirow{2}{*}{$0.0081$} & \textsc{IADR} & $\mathbf{19.99}$ & $\mathbf{2.07}$ & $\mathbf{0.64}$ \\
 & Specialist weight interpolation & $91.85$ & $15.48$ & $78.35$ \\
\bottomrule
\end{tabular}
\end{table}

\section{Reproducibility}
\label{app:repro}

\paragraph{Code release.}
The \textsc{IADR} implementation, baseline configurations, dataset
generation scripts, and bootstrap evaluation code are available at
\url{https://github.com/ArcueidCroft/IADR}. The README gives the
commands for generating data, training the models, and evaluating checkpoints.

\paragraph{Random seeds.}
All training runs reported in Table~\ref{tab:main} and
Table~\ref{tab:ablation} use a single fixed random seed per model;
seed values and the seed-dependent components (parameter
initialisation, mini-batch sampling, data-augmentation noise) are
recorded in the configuration files released with the code. Reported
metric variability comes from a non-parametric trajectory bootstrap
($B = 2000$) at evaluation time rather than from training-seed
resampling; the bootstrap procedure and its assumptions are described
in Appendix~\ref{app:exp-details:metrics}.

\paragraph{Runtime and hardware.}
All training was carried out on a single NVIDIA V100 GPU with $100$ epochs for DISCO and $20{,}000$ epochs for all other models. Per-run wall-clock times are: $16$\,h for \textsc{IADR},
$3$\,h for each of the eight per-$\varepsilon$ DR specialists,
$60$\,h for U-Net, $65$\,h for FNO, $72$\,h for DISCO, and
$18$\,h, $15$\,h, $14$\,h respectively for the Hyper-D, Hyper-R, and
Blind DR ablations, summing to roughly $284$ GPU-hours for the training runs listed here. Single-step inference timings on a $256 \times 256$ grid
are reported in Table~\ref{tab:step-time}.

\paragraph{Reference solver.}
Ground-truth trajectories on circles use the closed-form law
$R(t) = \sqrt{R_0^{2} - 2t}$. Ground-truth trajectories on
non-circular geometries (spiral disks and five-pointed stars) are
produced by the higher-order DR splitting solver of
Appendix~\ref{app:higher-order} at the same time step
$\delta t = 6.10 \times 10^{-5}$ as the trained models. The solver
configuration and dataset-construction details are in
Appendix~\ref{app:exp-details:datasets}.

\newpage
\section*{NeurIPS Paper Checklist}

\begin{enumerate}

\item {\bf Claims}
    \item[] Question: Do the main claims made in the abstract and introduction accurately reflect the paper's contributions and scope?
    \item[] Answer: \answerYes{}
    \item[] Justification: The three contributions listed in Section~\ref{sec:intro} match the slice-wise rollout result stated as Theorem~\ref{thm:rollout-main} in Section~\ref{sec:method:theory}, with its formal version and complete proof given as Theorem~\ref{thm:rollout-slice} in Appendix~\ref{app:proofs:slice} and a tighter linear non-expansive special case (Theorem~\ref{thm:rollout-linear-special}) in Appendix~\ref{app:proofs:linear}, together with the experimental results reported in Section~\ref{sec:exp}. The scope is limited to two-dimensional phase-field mean-curvature flow over the interface-width range $[\varepsilon_{\min},\varepsilon_{\max}]$ introduced in Section~\ref{sec:method:problem}.

\item {\bf Limitations}
    \item[] Question: Does the paper discuss the limitations of the work performed by the authors?
    \item[] Answer: \answerYes{}
    \item[] Justification: Section~\ref{sec:discussion} states the scope of the current validation and the planned extensions. Section~\ref{sec:method:theory} distinguishes the conditional rollout bound from the empirical long-horizon results.

\item {\bf Theory assumptions and proofs}
    \item[] Question: For each theoretical result, does the paper provide the full set of assumptions and a complete (and correct) proof?
    \item[] Answer: \answerYes{}
    \item[] Justification: Theorem~\ref{thm:rollout-main} (slice-wise informal) is stated in Section~\ref{sec:method:theory}; its formal version (Theorem~\ref{thm:rollout-slice}) and complete proof are given in Appendix~\ref{app:proofs:slice}, with its three direct hypotheses (Assumptions~\ref{assum:lip-T}--\ref{assum:consistency}) recorded in the same appendix. A tighter linear non-expansive special case is stated and proved as Theorem~\ref{thm:rollout-linear-special} in Appendix~\ref{app:proofs:linear}. The problem-level assumptions are formalised as Assumptions~\ref{assum:coverage}--\ref{assum:eps-lipschitz} in Appendix~\ref{app:proofs:slice}, and the slice-wise Lipschitz estimate used in the proof is recorded as Lemma~\ref{lem:dr-slice-lipschitz} in Appendix~\ref{app:proofs:slice-lipschitz}.

    \item {\bf Experimental result reproducibility}
    \item[] Question: Does the paper fully disclose all the information needed to reproduce the main experimental results of the paper to the extent that it affects the main claims and/or conclusions of the paper (regardless of whether the code and data are provided or not)?
    \item[] Answer: \answerYes{}
    \item[] Justification: The neural operator architecture, hypernetwork design, and training objective are described in Section~\ref{sec:method}; the training set, four evaluation tasks, time step, grid resolution, interface-width range, and reference solver are described in Section~\ref{sec:exp:setup}, with full per-shape sample counts and trajectory budgets given in Appendix~\ref{app:exp-details:datasets} (Table~\ref{tab:dataset-config}). Architecture and main training settings are described in Appendix~\ref{app:impl-details}, with further optimisation settings specified in the configuration files referenced there, and the discrete metric formulas used to score every model are given in Appendix~\ref{app:exp-details:metrics}.

\item {\bf Open access to data and code}
    \item[] Question: Does the paper provide open access to the data and code, with sufficient instructions to faithfully reproduce the main experimental results, as described in supplemental material?
    \item[] Answer: \answerYes{}
    \item[] Justification: The public repository at \url{https://github.com/ArcueidCroft/IADR} provides the \textsc{IADR} implementation, baseline configurations, dataset generation scripts, and bootstrap evaluation code. The README gives the commands for training and evaluation.

\item {\bf Experimental setting/details}
    \item[] Question: Does the paper specify all the training and test details (e.g., data splits, hyperparameters, how they were chosen, type of optimizer) necessary to understand the results?
    \item[] Answer: \answerYes{}
    \item[] Justification: Section~\ref{sec:exp:setup} states the training/evaluation split (training on circles at five widths $E_{\text{train}}$; evaluation on the four-task suite covering in-distribution, $\varepsilon$-OOD, geometry-OOD, and joint-OOD settings), the training budgets ($100$ epochs for DISCO and $20{,}000$ epochs for all other models), and the reference-solution provenance for each shape. Appendix~\ref{app:impl-details} provides architecture and training details for \textsc{IADR} and the baselines (target-network kernel sizes, hypernetwork depth/width, learning rate, and rollout-training setting), with further optimisation settings specified in the configuration files referenced there, and Appendix~\ref{app:exp-details:datasets} fixes the dataset construction (sample counts, trajectory budgets, initial-set parameters).

\item {\bf Experiment statistical significance}
    \item[] Question: Does the paper report error bars suitably and correctly defined or other appropriate information about the statistical significance of the experiments?
    \item[] Answer: \answerYes{}
    \item[] Justification: The errors in Tables~\ref{tab:main}, \ref{tab:main-full}, and~\ref{tab:ablation} are bootstrap means, with their associated $1$-sigma standard deviations reported alongside the means in Tables~\ref{tab:main-full} and~\ref{tab:ablation}. These statistics are computed from $B=2000$ bootstrap resamples of per-trajectory metric values and reflect trajectory-level sampling variability at a fixed model checkpoint, shape, and conditioning $\varepsilon$. Table~\ref{tab:step-time} reports the mean and standard deviation of single-step inference time across $\varepsilon$ within each shape.

\item {\bf Experiments compute resources}
    \item[] Question: For each experiment, does the paper provide sufficient information on the computer resources (type of compute workers, memory, time of execution) needed to reproduce the experiments?
    \item[] Answer: \answerYes{}
    \item[] Justification: All training runs were carried out on a single NVIDIA V100 GPU. The wall-clock time per run is $16$\,h for \textsc{IADR}, $3$\,h for each of the eight per-$\varepsilon$ DR specialists, $60$\,h for U-Net, $65$\,h for FNO, $72$\,h for DISCO, $18$\,h for Hyper-D, $15$\,h for Hyper-R, and $14$\,h for Blind DR, summing to roughly $284$ GPU-hours for the training runs listed here. The full research project, including preliminary architecture searches, baseline-tuning iterations, and failed runs not reported in the paper, consumed more than ten times this amount (above $2{,}900$ GPU-hours of total V100 compute).
    
\item {\bf Code of ethics}
    \item[] Question: Does the research conducted in the paper conform, in every respect, with the NeurIPS Code of Ethics \url{https://neurips.cc/public/EthicsGuidelines}?
    \item[] Answer: \answerYes{}
    \item[] Justification: The research conforms in every respect to the NeurIPS Code of Ethics. The work does not involve human subjects, personal data, or assets scraped from the web.

\item {\bf Broader impacts}
    \item[] Question: Does the paper discuss both potential positive societal impacts and negative societal impacts of the work performed?
    \item[] Answer: \answerNA{}
    \item[] Justification: The paper proposes a numerical surrogate for a geometric partial differential equation (phase-field mean-curvature flow). It has no direct application to decision-making, user-facing systems, or other downstream deployments, and we do not identify a specific positive or negative societal impact beyond its contribution to scientific computing.
    
\item {\bf Safeguards}
    \item[] Question: Does the paper describe safeguards that have been put in place for responsible release of data or models that have a high risk for misuse (e.g., pre-trained language models, image generators, or scraped datasets)?
    \item[] Answer: \answerNA{}
    \item[] Justification: The paper releases neither pretrained language models, image generators, nor scraped datasets. The released artifacts (training and inference code for a phase-field mean-curvature surrogate) pose no identifiable risk of misuse.

\item {\bf Licenses for existing assets}
    \item[] Question: Are the creators or original owners of assets (e.g., code, data, models), used in the paper, properly credited and are the license and terms of use explicitly mentioned and properly respected?
    \item[] Answer: \answerYes{}
    \item[] Justification: We adapt the official DISCO codebase~\citep{morel2025disco} as the meta-learning baseline, retaining DISCO's own U-Net target and adapting the dataloader to the phase-field trajectories. Our DR splitting baseline is implemented following the description and reference code of~\citet{bretin2022jcp}; the FNO baseline~\citep{li2021fno} is re-implemented from the published paper. These works are credited at first use in Sections~\ref{sec:related}--\ref{sec:exp:setup}, and we respect the licenses of the upstream repositories. The repository README documents the installation of the upstream DISCO dependency. The four evaluation datasets (circles, spiral disks, five-pointed stars, and dumbbells) are constructed from scratch and introduce no third-party data.

\item {\bf New assets}
    \item[] Question: Are new assets introduced in the paper well documented and is the documentation provided alongside the assets?
    \item[] Answer: \answerYes{}
    \item[] Justification: We release the \textsc{IADR} training and evaluation code together with the dataset generation scripts (circles, spiral disks, and five-pointed stars at the interface-width range $E$, with analytic reference trajectories for circles and higher-order DR splitting reference trajectories for spiral disks and five-pointed stars). The repository linked in Q5 documents the data generation, training, and evaluation commands in its README and provides the configurations described in Appendix~\ref{app:impl-details}. Trained model checkpoints are not released; the released scripts are sufficient to reproduce them under the protocol of Section~\ref{sec:exp:setup}.

\item {\bf Crowdsourcing and research with human subjects}
    \item[] Question: For crowdsourcing experiments and research with human subjects, does the paper include the full text of instructions given to participants and screenshots, if applicable, as well as details about compensation (if any)?
    \item[] Answer: \answerNA{}
    \item[] Justification: The research does not involve crowdsourcing or human subjects.

\item {\bf Institutional review board (IRB) approvals or equivalent for research with human subjects}
    \item[] Question: Does the paper describe potential risks incurred by study participants, whether such risks were disclosed to the subjects, and whether Institutional Review Board (IRB) approvals (or an equivalent approval/review based on the requirements of your country or institution) were obtained?
    \item[] Answer: \answerNA{}
    \item[] Justification: The research does not involve human subjects; no IRB approval is required.

\item {\bf Declaration of LLM usage}
    \item[] Question: Does the paper describe the usage of LLMs if it is an important, original, or non-standard component of the core methods in this research? Note that if the LLM is used only for writing, editing, or formatting purposes and does \emph{not} impact the core methodology, scientific rigor, or originality of the research, declaration is not required.
    \item[] Answer: \answerNA{}
    \item[] Justification: Large language models did not contribute to the core methodology, the experimental design, the theoretical analysis, or any algorithmic component of this research. Any LLM usage was limited to incidental writing and editing assistance, which falls outside the scope of this declaration.

\end{enumerate}

\end{document}